%% file: 0-main.tex
\documentclass{article}
\pdfoutput=1
\PassOptionsToPackage{numbers, compress}{natbib}

\usepackage[preprint]{neurips_2019}

\input{0-preamble.tex}

\newcommand\boldhead[1]{\vspace{0.03in}\noindent\textbf{#1: }}
\newcommand{\interpfig}[1]{\includegraphics[width=1.2cm,height=1.2cm]{#1}}
\newcommand{\interpfigbbc}[1]{\includegraphics[width=1.2cm,height=1.2cm]{#1}}
\newcommand{\interpfigkth}[1]{\includegraphics[width=0.94cm,height=0.94cm]{#1}}
\newcommand{\interpfigbair}[1]{\includegraphics[width=1.1cm,height=1.1cm]{#1}}
\newcommand{\interpfiglarge}[1]{\includegraphics[width=1.6cm,height=1.6cm]{#1}}

\title{Video Interpolation and Prediction with Unsupervised Landmarks}

\author{%
  Kevin J. Shih\thanks{Equal contribution} \And Aysegul Dundar\footnotemark[1] \And Animesh Garg \And Robert Pottorf \And Andrew Tao \And Bryan Catanzaro\\  
  NVIDIA Corporation
}

\begin{document}

\maketitle

\begin{abstract}

Prediction and interpolation for long-range video data involves the complex task of modeling motion
trajectories for each visible object, occlusions and dis-occlusions, as well as appearance
changes due to viewpoint and lighting. Optical flow based techniques
generalize but are suitable only for short temporal ranges. Many
methods opt to project the video frames to a low dimensional latent
space, achieving long-range predictions. However, these latent
representations are often non-interpretable, and therefore difficult
to manipulate. 
This work poses video prediction and interpolation as unsupervised latent structure inference followed by a temporal prediction in this latent space. The latent representations capture foreground semantics without explicit supervision such as keypoints or poses. 
Further, as each landmark can be mapped to a coordinate indicating where a semantic part is positioned, we can reliably interpolate within the coordinate domain to achieve predictable motion interpolation. Given an image decoder capable of mapping these landmarks
back to the image domain, we are able to achieve high-quality
long-range video interpolation and extrapolation by operating on the
landmark representation space.

\end{abstract}

\input{1-sec_introduction}
\input{2-sec_method}

\input{3-sec_experiments}

\input{4-sec_related_work}
\input{5-sec_conclusion.tex}
\clearpage
{
\bibliography{vid-pred}
\bibliographystyle{ieee}
}
\clearpage
\input{6-appendix.tex}

\end{document}

%% file: 0-preamble.tex
\usepackage{graphics}
\usepackage{algorithm}
\usepackage{algpseudocode}
\usepackage{pgfplots}
\usepackage{pgfplotstable}
\usepackage{wrapfig}
\DeclareGraphicsExtensions{.pdf,.png,.jpg}

\usepackage{xcolor}
\definecolor{forestgreen}{rgb}{0.13, 0.55, 0.13}
\definecolor{fulvous}{rgb}{0.86, 0.52, 0.0}
\definecolor{glaucous}{rgb}{0.38, 0.51, 0.71}
\definecolor{lava}{rgb}{0.81, 0.06, 0.13}
\definecolor{chromeyellow}{rgb}{1.0, 0.65, 0.0}
\definecolor{brightube}{rgb}{0.82, 0.62, 0.91}

\usepackage[skip=2pt,font=small]{caption}
\usepackage{subcaption}
\usepackage[rightcaption]{sidecap}
\usepackage{pbox}

\usepackage{mathtools}
\usepackage{amsmath, amssymb, amscd}
\usepackage{ wasysym } %
\usepackage{amsfonts}
\usepackage{mathptmx} %
\usepackage{gensymb} 
\usepackage{nicefrac}       %

\DeclareMathAlphabet{\mathcal}{OMS}{lmsy}{m}{n}
\DeclareSymbolFont{largesymbols}{OMX}{cmex}{m}{n}
\usepackage{textcomp} %

\usepackage{algorithm} %
\usepackage{algorithmicx, algpseudocode} %

\usepackage{array} %
\usepackage{tabularx}
\usepackage{multirow}
\usepackage{multicol}
\usepackage{booktabs}

\usepackage[utf8]{inputenc} %
\usepackage[T1]{fontenc}    %
\usepackage[english]{babel} %
\usepackage{units}
\usepackage{bm}
\usepackage{times} %
\usepackage{xspace}
\usepackage{flushend}%
\usepackage{csquotes}
\usepackage{makeidx}
\usepackage{blindtext}

\usepackage{enumitem}

\usepackage{soul} %
\usepackage{subfiles} %

\usepackage[yyyymmdd]{datetime}

\date{\protect\formatdate{1}{1}{2001}}

\usepackage{url}
\makeatletter
\g@addto@macro{\UrlBreaks}{\UrlOrds}
\makeatother
\usepackage{color}

\usepackage{marginnote}
\usepackage{soul} %

\usepackage[colorinlistoftodos]{todonotes}

\newcommand{\ignore}[1]{}



\pgfplotstableread[col sep=comma]{data/kth_prediction_perceptual.txt}\kthpreddata
\pgfplotstabletranspose[input colnames to=t,colnames from=t]\kthpredtpose\kthpreddata

\pgfplotstableread[col sep=comma]{data/kth_ssim_psnr.txt}\kthpreddata
\pgfplotstabletranspose[input colnames to=t,colnames from=t]\kthpredtpose\kthpreddata

\pgfplotstableread[col sep=comma]{data/bair_prediction.txt}\bairpreddata
\pgfplotstabletranspose[input colnames to=t,colnames from=t]\bairpredtpose\bairpreddata

\pgfplotstableread[col sep=comma]{data/bair_prediction_ssim.txt}\bairpreddatassim
\pgfplotstabletranspose[input colnames to=t,colnames from=t]\bairpredtposessim\bairpreddatassim
\pgfplotstableread[col sep=comma]{data/bair_prediction_psnr.txt}\bairpreddatapsnr
\pgfplotstabletranspose[input colnames to=t,colnames from=t]\bairpredtposepsnr\bairpreddatapsnr

%% file: 1-sec_introduction.tex
\section{Introduction}
Modeling long-range video data involves the complex task of handling motion
trajectories for each visible object, occlusions and dis-occlusions, as well appearance
changes due to viewpoint and lighting. Common applications include video
interpolation and video prediction, which involve inferring
unseen frames conditioned on neighboring, or past frames respectively. While short term prediction performance can be achieved through variety of methods, long-term modelling remains an ongoing challenge.
We propose a \textit{self-supervised} pipeline to model \textit{long-term video
dynamics} in a \textit{latent, explicit} pose representation, requiring only
rough object-level bounding boxes for training.

Existing methods can be roughly categorized into two camps: (a) flow-based methods that infer pixel-level linear transformations between frames, and (b) hidden state methods that infer a latent hidden space with possibly non-linear dynamics. While flow-based methods can produce sharp results, they only work well over short
temporal ranges and cannot synthesize novel, unseen pixels produced during dis-occlusion. 
Hidden state approaches on the other hand are capable of longer-range inferences, but tend be domain
specific and produce non-interpretable latent encodings, and consequently are difficult to regularize and explicitly manipulate.

In this work, we consider learning a latent representation of $K$ landmark ``keypoints," each describing the spatially local appearance and 2D coordinates of a single rigid component. This representation is interpretable, and follows naturally when modeling the dynamics of linked or highly correlated rigid objects in a scene. 
A straightforward approach to learning object keypoints would be to
manually label thousands of images before training a supervised model. Although use of supervised keypoints has been explored in prior work~\cite{villegas2017learning}, creating large-scale accurate keypoint-level annotation is both difficult and expensive, considering the diversity of
everyday objects. 

Our work builds upon existing methods for $\textit{unsupervised}$ landmark identification~\cite{jakab2018unsupervised,lorenz2019unsupervised}. These methods learn localized landmarks that activate consistently on the same semantic locations, independent of the class of typical transformations. We extend this body of work by exploring the use of unsupervised landmarks for long-range video frame prediction and interpolation. We demonstrate that a constrained and parameterized representation is both sufficiently expressive and stable to achieve long-range quality estimates for the video prediction task, while also providing intuitive results during interpolation tasks. Our main contributions are as follows:
\begin{enumerate}[
    topsep=0pt,
    noitemsep,
    partopsep=1ex,
    parsep=0.5ex,
    leftmargin=*,
    itemindent=3ex
    ]
\item We propose an unsupervised video modeling method that can interpolate and extrapolate over 100 frames into the future while maintaining the structure of the moving foreground object. 
\item We improve upon recent work in unsupervised landmark learning using conditional image reconstruction.
\item We demonstrate how to safely manipulate pose representations based on 2D Gaussian heatmaps in both predictive and interpolative settings.%
\end{enumerate}

%% file: 2-sec_method.tex
\begin{figure}
  \includegraphics[width=\textwidth,trim={0, 5cm, 0, 1.9cm}, clip]{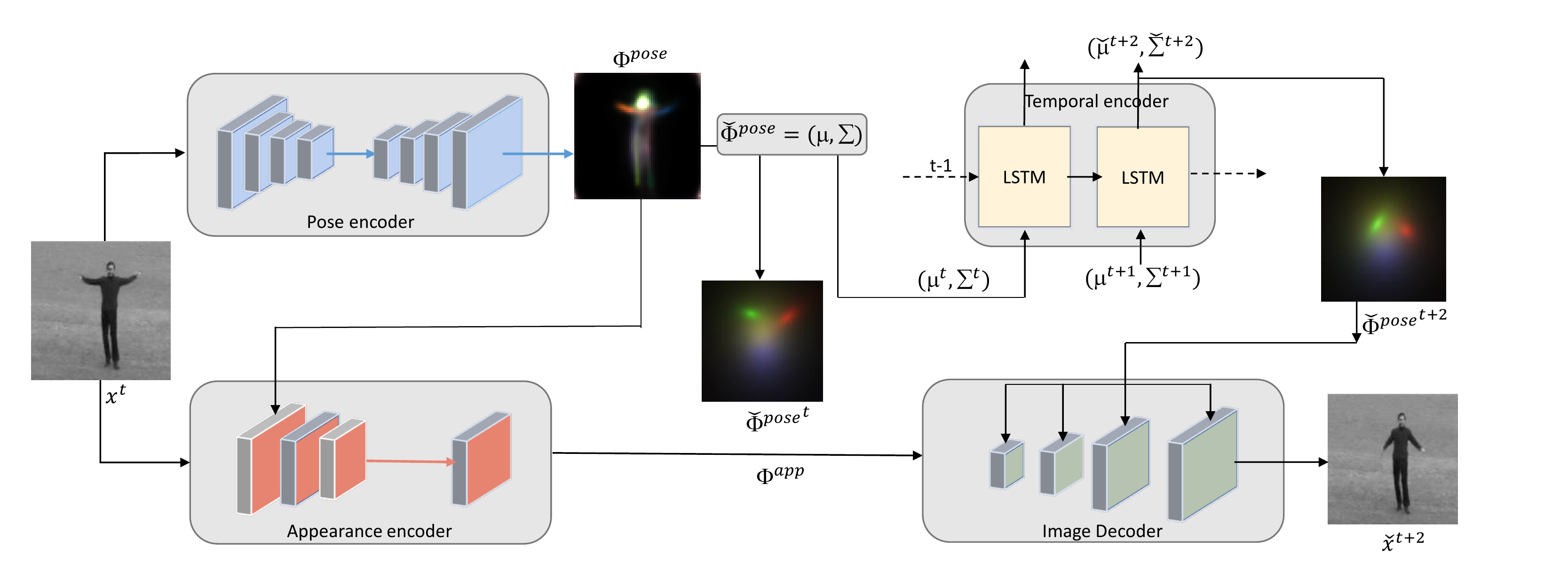}
\caption{Pose and appearance encoders project the image into the factorized pose-appearance space. 2D Gaussian is fit to each activation in the pose code. Only a subset of the pose landmarks in $\Phi^{pose^t}$ are visualized for clarity.
Gaussian parameters are fed to an LSTM after covariance matrixes go through Cholesky decomposition which predicts the next pose represenation in Gaussian parameter space (${\mu_k}^{t+2}, {\Sigma_k}^{t+2}$).
$\Phi^{pose^{t+2}}$ is reconstructed from these parameters and is used to reconstruct the future frame with the $\Phi^{app}$.}
\label{fig:method}
\end{figure}

\section{Method}
Our method can be broken down into three primary components: an image
encoder that projects the image into the factorized
pose-appearance space, an image decoder that reconstructs the image
from the pose-appearance space, and a temporal dynamics component that
predicts how the pose representation changes through time. Overview of the method is shown in Fig. \ref{fig:method}.

\subsection{Unsupervised Image Factorization}
Our method for learning our pose and appearance factorization is based on the unsupervised landmarks learning work proposed by Lorenz et al.~\cite{lorenz2019unsupervised}. The method separates an input image into an appearance-invariant pose representation and a pose-invariant appearance representation, then attempts to reconstruct the original input image in order to learn landmarks. The network for this method is built from three models: an image to pose encoder, an image to appearance encoder, and an image decoder that attempts to reconstruct the image from the factorized pose and appearance representations.

The goal of the pose encoder $\Phi^{pose} = \mathit{Enc}^{pose}(x)$ is to take an input image $x$ and output a set of consistent part activation maps. Critically, we want these part activation maps to be invariant to changes in local appearance, as well as to be consistent across deformations. For example, a heatmap that activates on a person's right hand should be invariant across varying skin tones and lighting conditions, as well track the right hand's location across varying deformations and translations.

The appearance encoder $\Phi^{app} = \mathit{Enc}^{app}(x; \mathit{Enc}^{pose}(x))$ extracts local appearance information, conditioned on the pose-encoder's activation maps. Given an input image $x$, the pose encoder will first provide  $K \times H\times W$ part activation maps $\Phi^{pose}$. To extract local appearance vectors, the appearance encoder  projects the image to a $C\times H\times W$ appearance feature map $M^{app}$. We compute the appearance vector for the $k$th pose activation map as:
\begin{equation}
  \Phi_{k,c}^{app} = \sum_{i}^{H} \sum_{j}^{W} \Phi^{pose}_{k,i,j} M^{app}_{c,i,j} \text{  for } c = 1...C,
\end{equation}
, giving us $K$ $C$-dimensional appearance vectors. Here, each activation map in $\Phi^{pose}$ is softmax-normalized.

Finally, the image decoder attempts to reconstruct the original input image by combining the pose information from the $K$ activation maps with the pooled appearance vectors for each of the $K$ parts. As in~\cite{lorenz2019unsupervised}, instead of conditioning on the raw softmax-normalized activation pose maps, we instead fit a 2D Gaussian to each activation of the $K$ activation maps by estimating their respective means and covariance matrices. Each part is represented by $\widetilde{\Phi}^{pose}_k = (\mu_k, \Sigma_k) $, where $\mu_k \in \mathbb{R}^2$ and $\Sigma_k \in \mathbb{R}^{2\times 2}$. The 2D Gaussian approximation is critical to our ability to manipulate our pose representation as it forces each part activation map into a unimodal representation with a simple parameterization.

The pose encoder, shape encoder, and image decoder are jointly trained in a fully self-supervised fashion, using the final image reconstruction task as the only supervision. Again, we roughly follow the training method as detailed in ~\cite{lorenz2019unsupervised}, with some differences in choice of loss functions.

Color jittering and thin-plate-spline (TPS) warping are incorporated into the training pipeline to enforce the pose encoder's appearance invariance and localization properties respectively. Let $T_{cj}(x)$ and $T_{tps}(x)$ represent color jittering and thin-plate spline warping operations on the input image $x$ respectively. Our training procedure can be expressed as follows:
\begin{align}
  &\Phi^{pose} = \mathit{Enc}^{pose}(T_{cj}(x))\\
  &\Phi^{app} = \mathit{Enc}^{app}(T_{tps}(x); \mathit{Enc}^{pose}(T_{tps}(x)))\\
  &\tilde{x} = \mathit{Dec}(\widetilde{\Phi}^{pose}, \Phi^{app}),\label{eq:dec_out}
\end{align}
, where the goal is to minimize the reconstruction loss between the original input $x$ and the reconstruction $\tilde{x}$. As can be seen, neither the shape encoder nor the appearance encoder are ever given access to the un-perturbed target $x$. The pose information feeding into the image decoder $\mathit{Dec}(\cdot, \cdot)$ is based on the color-jittered input image, where only the local appearance information is perturbed. The appearance information is computed from a thin-plate-spline warped version of the input image, where the pose information is perturbed. Furthermore, as the shape encoder is also executed on the TPS-warped input image, it needs to localize the same corresponding locations in their new warped coordinates in order for the correct appearance information to be captured. 

As presented in~\cite{jakab2018unsupervised,lorenz2019unsupervised},
we also use the existing motion in the video data as a source of pose
perturbation. Specifically, in addition to the $T_{cj}(x)$ -
$T_{tps}(x)$ pairing, we also include an additional pairing where the
$T_{tps}$ is replaced by sampling a random video frame between 3 and
60 frames away from the current frame $x$ (referred to as
$T_{temp}(x)$). 

\subsection{Image Decoder and Reconstruction Losses}
Our image decoder and losses differ from those of Lorenz et al.~\cite{lorenz2019unsupervised}. Our image decoder is inspired from the architecture proposed in SPADE~\cite{park2019semantic}. In SPADE, semantic maps are used to predict spatially-aware affine transformation parameters for normalization schemes such as InstanceNorm. Herein, we project the 2D Gaussian parameters from $\widetilde{\Phi}^{pose}$ to a heatmap of the target output width and height to use as semantic maps in the SPADE architecture. Following ~\cite{lorenz2019unsupervised}, we use the formula:
\begin{equation}
  s(k,l) = \frac{1}{1+(l - \mu_k)^\mathsf{T}\Sigma_k^{-1}(l - \mu_k)}
\end{equation}
, where $s(k,l)$ is the heatmap value for part map $k$ at coordinate location $l$.

For losses, we use a VGG perceptual-loss similar to the one used by Jakab et al.~\cite{jakab2018unsupervised} with pre-trained ImageNet weights, combined with an adversarial loss to improve realism and a standard pixel-level mean square error loss. Our reconstruction loss is expressed as follows:
\begin{equation}
  \mathcal{L}_{recon}(x, \tilde{x}) = \lambda_{vgg} \mathcal{L}_{vgg}(x, \tilde{x}) + \lambda_{MSE}\textit{MSE}(x,\tilde{x}) + \mathcal{L}_{adv}(\tilde{x}).
\end{equation}
Detailed formulations are available in the appendix.

\subsection{Interpolating Unsupervised Pose Representations}
Since our unsupervised landmarks-based pose representation can be explicitly interpreted as
localized part labels, it is natural to manipulate and
interpolate within this space. 
However, we need to be careful to ensure that the covariance matrix for each landmark remains positive definite throughout any manipulation. This can be done in many parameterizations: directly in the covariance parameters, or indirectly in the parameters of the Cholesky decomposition, or the angle and magnitude of the principal components, or even along an optimal transport as a Wasserstein Barycenter~\cite{chen2019optimal}. 
It is worth noting that owing to the wide variety of possible projected deformations, none of these methods dominate in all conditions. 

In this work, we utilize the parameters of the Cholesky decomposition because they are both easy to compute, and guarantee covariance validity in both interpolation and extrapolation paradigms. The Cholesky decomposition of each covariance matrix is the unique matrix $L_k$ with positive diagonal such that $\Sigma_k = L_kL_k^{\mathsf{T}}$ where $L_k$ is a lower-triangular. In the case of our 2-dimensional keypoints, this results in three scalars (two positive, one unconstrained) which uniquely parameterize the covariance of each keypoint. When coupled with two dimensional mean $\mu_k$, these five parameters form a state vector for keypoint $k$ at timestep $t$. Interpolation is done linearly:

\begin{equation}
  \begin{bmatrix} \hat{\mu}_k \\ \hat{L}_k \end{bmatrix}_{t = \alpha} = (1-\alpha)\begin{bmatrix} \mu_k \\ L_k \end{bmatrix}_{t=0} + \alpha\begin{bmatrix} \mu_k \\ L_k \end{bmatrix}_{t=1}.
\end{equation}

\subsection{Extrapolating Unsupervised Pose Representations}
\begin{wrapfigure}{r}{0.3\textwidth}
    \centering
    \vspace{-15pt}
  \includegraphics[scale=0.5,trim=125 225 605 60, clip]{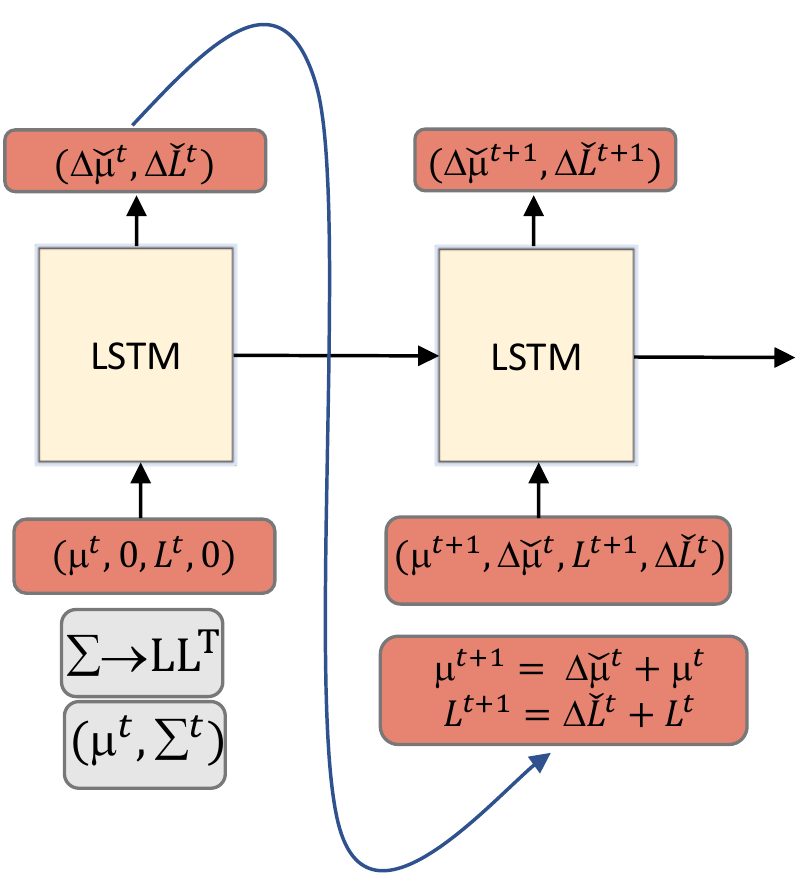}
  \caption{The LSTM predicts perturbations to the Gaussian means and Cholesky parameters of the covariances.}
  \label{fig:lstm}
  \vspace{-12pt}
\end{wrapfigure}
We use an LSTM to model the non-linear temporal dynamics of the pose representation~\cite{hochreiter1997lstm}.
As in the case of interpolation, care must be taken to maintain positive definite covariance matrices during prediction. Thus we use the parameters of the Cholesky decomposition of the covariance matrices as the prediction targets.
In practice, the LSTM may produce an estimate for $L$ which is not a valid Cholesky factor, but even this case will still produce a valid covariance matrix when $LL^T$ is computed.

An illustration of our LSTM setup is depicted in Fig.\ref{fig:lstm}. In addition to the Cholesky parameterization, we also found that predicting the residual during extrapolation, rather than the state directly, was important to robust long term predictions. This state residual (aka error state in control literature~\cite{madyasthaekf}) is well known to improve estimation performance in a variety of tasks with non-linear dynamics. In addition to maintaining an unconstrained gradient path stretching far into the past, the error-state also induces our model to learn local deformations of the state, which are easier to learn as a result of being small, and nearly linear. In practice. this helped improve both training performance and generalization. At each time-step, the LSTM takes a concatenation of the previous state and state residual to predict the next state residual.

%% file: 3-sec_experiments.tex
\section{Experiments}
Our experiments evaluate the interpolability of our
unsupervised pose representation, as well as our model's ability to
extrapolate far into the future. 

\noindent \textbf{Datasets.} We test our method on three video sequence datasets:
BBC Pose~\cite{Charles13}, BAIR Push ~\cite{finn2016unsupervised}, and KTH Action~\cite{laptev2004recognizing} datasets.
For each dataset, we first train our encoders and decoder on individual image frames. For video prediction, we separately project training video sequences into the latent pose representation, then train an LSTM to model the temporal dynamics of the pose representation across time, one per dataset, and do not condition on action labels. Please refer to Appendix for further details.

\noindent \textbf{Metrics.} Evaluating the quality of long sequences is inherently difficult, as the number of valid sequences
grows exponentially with length, and the reference is only one of many valid outcomes. While we include image quality metrics such as
PSNR and SSIM for completeness, we note that their relevance decreases as the predicted trajectory diverges from the reference. 

We also include a metric based on the cosine distance of pre-trained CNN features, which has been found to correlate better with human perception than SSIM and PSNR. We use the LPIPS v0.1 metric implementation provided by
\cite{zhang2018perceptual,LPIPSGithub}. 
Because we are already using the VGG model for our training
loss, we report the LPIPS metric using both VGG and AlexNet features. Further, we are interested in how well each model maintains the videos' structural integrity over long predictions, thus we use the LPIPS metrics to identify worst-case video sequences to compare qualitatively.

\subsection{BBC Pose Dataset}
The BBC Pose dataset contains 20 video sequences featuring 9 unique
sign language interpreters. Individual frames are annotated with
keypoint annotations for the signer. The train and validation splits
feature 5 of the signers across 5 train and 5 validation videos.
For inference, we initialize our LSTM with 2 frames and use the appearance information from the
2nd for the entire sequence.

\begin{table}
 \centering
 \vspace{-10pt}
 \caption{Evaluation unsupervised pose
    representation on the BBC keypoint evaluation task.
    Our implementation
    outperforms that of ~\cite{lorenz2019unsupervised} by incorporating SPADE normalization.
    }
  \begin{tabular}{ll|cc}
    \multicolumn{2}{c|}{BBC Pose} & accuracy & best\\
    \hline
    supervised & Charles\cite{Charles13} & 79.9\%\\
    & Pfister\cite{pfister2015flowing} & 88.0\%\\
    \hline
    unsupervised & Jakab\cite{jakab2018unsupervised} & 68.4\%\\
    & Lorenz\cite{lorenz2019unsupervised} & 74.5\%\\
    & Ours (Lorenz) & $74.2\% \pm 1.6$ & 75.7\%\\
    & Ours (+spade) & $75.0\% \pm 0.9$ & 76.1\%\\
  \end{tabular}
  \label{tbl:bbc_supervised}
  \vspace{-0.3cm}
\end{table}

\boldhead{Landmark Quality Evaluation with Supervised Keypoints}
We first evaluate our unsupervised landmark quality on the BBC Pose annotated keypoints. As with prior works such as \cite{jakab2018unsupervised} and our target reproduction \cite{lorenz2019unsupervised}, we fit a linear regressor to our learned landmark locations from our pose representation to supervised keypoint coordinates.
Following \cite{jakab2018unsupervised} and \cite{lorenz2019unsupervised}, we first roughly crop around the presenter in the train and test images using the provided annotations. Additional details included in the appendix. A comparison of our keypoint prediction accuracy is shown in
Table~\ref{tbl:bbc_supervised}. Keypoint accuracy is reported where a correct prediction must be within six pixels
of the target. While our work closely follows the method proposed
in~\cite{lorenz2019unsupervised} for learning unsupervised landmarks
as our pose representation, our specific implementation differs in our
choice of loss functions, image decoder algorithm, and minor architectural
details for our pose and appearance encoders $\mathit{Enc}^{pose}$ and
$\mathit{Enc}^{app}$. We implement two variants: with and without
SPADE~\cite{park2019semantic} and conduct 3 runs of each. For the
without variant, we simply include the landmark heatmaps as input channels. We
demonstrate that using the landmark heatmaps as semantic maps to
modulate the normalization affine parameters improves overall landmark quality.

\begin{figure}
  \centering
  \setlength\tabcolsep{0.32pt}
  \renewcommand{\arraystretch}{0.15}
  \begin{tabular}{cccccccccccc}
    &t=1  &t=10 &t=20 &t=30 &t=40 &t=50 &t=60 &t=70 &t=80 &t=90 &t=100\\\\
    
    \rotatebox{90}{~~~GT}&
     \interpfigbbc{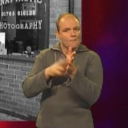}&
     \interpfigbbc{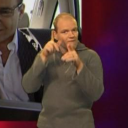}&
     \interpfigbbc{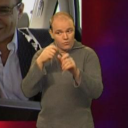}&
     \interpfigbbc{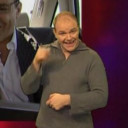}&
     \interpfigbbc{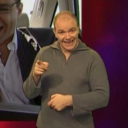}&
     \interpfigbbc{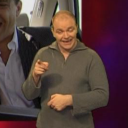}&
     \interpfigbbc{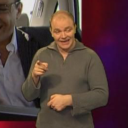}&
    \interpfigbbc{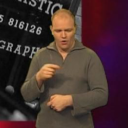}&
    \interpfigbbc{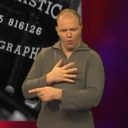}&
     \interpfigbbc{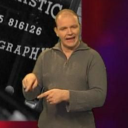}&
     \interpfigbbc{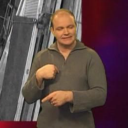}\\
    \rotatebox{90}{~~~Ours}&
    \interpfigbbc{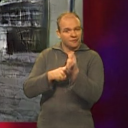}&
     \interpfigbbc{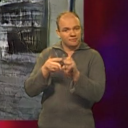}&
     \interpfigbbc{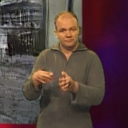}&
     \interpfigbbc{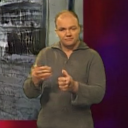}&
     \interpfigbbc{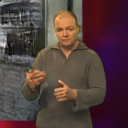}&
     \interpfigbbc{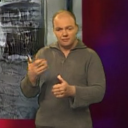}&
     \interpfigbbc{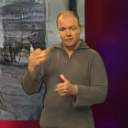}&
    \interpfigbbc{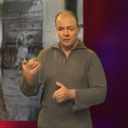}&
    \interpfigbbc{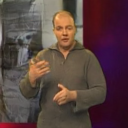}&
    \interpfigbbc{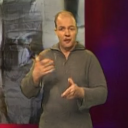}&
     \interpfigbbc{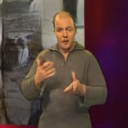}\\
    \rotatebox{90}{~~~GT}&
     \interpfigbbc{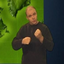}&
     \interpfigbbc{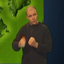}&
     \interpfigbbc{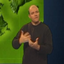}&
     \interpfigbbc{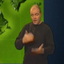}&
     \interpfigbbc{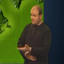}&
     \interpfigbbc{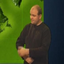}&
    \interpfigbbc{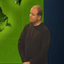}&
     \interpfigbbc{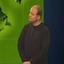}&
     \interpfigbbc{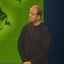}&
    \interpfigbbc{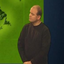}&
    \interpfigbbc{Video_pred/BBC_val_vid13/GT/000100.png}\\
    \rotatebox{90}{~~~Ours}&
     \interpfigbbc{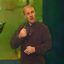}&
     \interpfigbbc{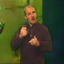}&
     \interpfigbbc{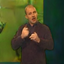}&
     \interpfigbbc{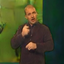}&
     \interpfigbbc{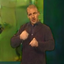}&
     \interpfigbbc{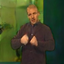}&
    \interpfigbbc{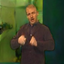}&
     \interpfigbbc{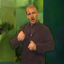}&
     \interpfigbbc{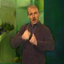}&
     \interpfigbbc{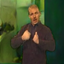}&
    \interpfigbbc{Video_pred/BBC_val_vid13/Ours/000100.png}\\
    \end{tabular}
  \caption{Qualitative results for long range video prediction. The first sequence features held out frames from the training set, while the second features a held out frames from the validation set (same person as first row, unseen attire). Our pose encoding generalizes well, but our generation pipeline struggles to render novel appearances, as seen in the second sequence where the signer's attire does not match that of the reference.}
  \label{Tbl:bbc_pred_images}
\end{figure}

\boldhead{Video Prediction} 
We qualitatively evaluate our video prediction results on the BBC Pose
dataset, conditioning on two held-out frames and generating the next
98. Qualitative results are shown in
Fig.\ref{Tbl:bbc_pred_images} on frames from the train and val splits. Additional video sequences are included in the supplemental material. The first sequence features held out frames from the train set, while the second (green background) features the same signer from the first row but wearing held-out attire. From the first sequence, we see that our model is able to extrapolate complex long-range hand and head gestures from the sign interpreters. In the second sequence, the predicted poses smoothly transition from the input, but the rendered signer, while still realistic, does not match the appearance of the ground truth. This demonstrates that our pose extraction
generalizes well across novel appearances, but the image generation
part of the pipeline requires significantly more variety in the
training data to accurately reproduce novel appearances. A possible
line of future work would be to incorporate techniques from few and
zero-shot appearance transfer to our generator pipeline.

\subsection{BAIR Action-Free Robot Pushing Dataset}
We evaluate our video prediction quality and interpolation quality on
the BAIR push dataset. This dataset features 64$\times$64 video frames
of a robot arm pushing a diverse set of objects around. 
\begin{wrapfigure}{r}{0.48\textwidth}
  \setlength\tabcolsep{0.32pt}
  \renewcommand{\arraystretch}{0.12}
  \begin{tabular}{cccccc}
    &t=2  & t=8 & t=16 & t=24 & t=30 \\
    \rotatebox{90}{~~~GT}&
     \interpfigbair{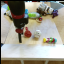}&
     \interpfigbair{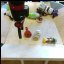}&
     \interpfigbair{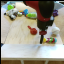}&
     \interpfigbair{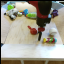}&
     \interpfigbair{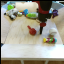}\\
          & Ours &
      \interpfigbair{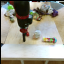}&
     \interpfigbair{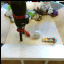}&
     \interpfigbair{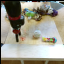}&
     \interpfigbair{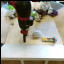}\\
       & SVGLP &
     \interpfigbair{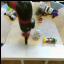}&
     \interpfigbair{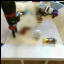}&
     \interpfigbair{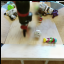}&
     \interpfigbair{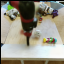}\\
       & SAVP Det&
     \interpfigbair{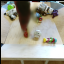}&
     \interpfigbair{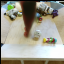}&
     \interpfigbair{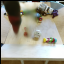}&
     \interpfigbair{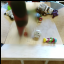}\\ 
    & SAVP &
     \interpfigbair{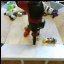}&
     \interpfigbair{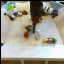}&
     \interpfigbair{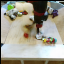}&
     \interpfigbair{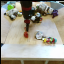}\\ 
  \end{tabular}
  \caption{Predicted frames from one of the test sequences that all models had a bad LPIPS score on. While other models suffer some blurring in the objects, our foreground objects remain sharp.}
  \label{fig:bair-failure-analysis}
  \vspace{-2em}
\end{wrapfigure}

\boldhead{Video Prediction}
For inference in the video prediction task, we follow the procedure of \cite{lee2018stochastic},
initializing our LSTM with the first two frames on each test sequence
and predicting the remaining 28. We compare against the SAVP
\cite{lee2018stochastic} and SVG-LP\cite{denton2018stochastic} using both VGG and AlexNet variants of the
LPIPS\cite{zhang2018perceptual} perceptual metric. However, because we
are not evaluating the benefits of stochasticity in this work, we
generate only a single sample from both SAVP and SVG-LP. We also
include the deterministic baseline from~\cite{lee2018stochastic} as our
primary comparison.

\begin{figure}
  \centering
  \begin{tikzpicture}[thick, scale=0.8]
    \begin{axis}[    name=ax1,
    ylabel = Mean LPIPS,
    ylabel near ticks,
    xlabel near ticks,
    title = LPIPS(VGG) on BAIR,
    legend columns=6,
    xtick distance=5,
    enlargelimits=false,
    y tick label style={
    /pgf/number format/.cd,
        fixed,
        fixed zerofill,
        precision=2,
    /tikz/.cd
    },
    width=\textwidth * 0.5,
    height=5cm,
    ymajorgrids,
    legend columns=1,
    legend entries={ours, SAVP-Det \cite{lee2018stochastic}, SAVP \cite{lee2018stochastic}, SVGLP \cite{denton2018stochastic}},
    legend style={at={(2.6,1)},anchor=north}
      ]
      \addplot [line width=0.5mm, color=glaucous,error bars/.cd,y dir=both, y explicit] table [x=t, y=oursvggmean, col sep=comma, y error=oursvggstderr] {\bairpredtpose};
      \addplot [line width=0.5mm, color=lava,error bars/.cd,y dir=both, y explicit] table [x=t, y=savpdetvggmean, col sep=comma, y error=savpdetvggstderr]{\bairpredtpose};
      \addplot [line width=0.5mm, color=black,error bars/.cd,y dir=both, y explicit] table [x=t, y=savp0vggmean, col sep=comma,y error=savp0vggstderr]{\bairpredtpose};
      \addplot [line width=0.5mm, color=forestgreen,error bars/.cd,y dir=both, y explicit] table [x=t, y=svgvggmean, col sep=comma,y error=svgvggstderr]{\bairpredtpose};
    \end{axis}
        \begin{axis}[    at={(ax1.south west)},
            yshift=-4.5cm,
    xlabel = Frame Number,
    ylabel = Mean LPIPS,
    ylabel near ticks,
    xlabel near ticks,
    title = LPIPS(Alex) on BAIR,
    legend columns=6,
    xtick distance=5,
    enlargelimits=false,
    y tick label style={
    /pgf/number format/.cd,
        fixed,
        fixed zerofill,
        precision=2,
    /tikz/.cd
    },
    width=\textwidth * 0.5,
    height=5cm,
    ymajorgrids
      ]
      \addplot [line width=0.5mm, color=glaucous,error bars/.cd,y dir=both, y explicit] table [x=t, y=oursalexmean, col sep=comma, y error=oursalexstderr] {\bairpredtpose};
      \addplot [line width=0.5mm, color=lava,error bars/.cd,y dir=both, y explicit] table [x=t, y=savpdetalexmean, col sep=comma, y error=savpdetalexstderr]{\bairpredtpose};
      \addplot [line width=0.5mm, color=black,error bars/.cd,y dir=both, y explicit] table [x=t, y=savp1alexmean, col sep=comma,y error=savp1alexstderr]{\bairpredtpose};
      \addplot [line width=0.5mm, color=forestgreen,error bars/.cd,y dir=both, y explicit] table [x=t, y=svgalexmean, col sep=comma,y error=svgalexstderr]{\bairpredtpose};
    \end{axis}
        \begin{axis}[    at={(ax1.south east)},
            name=axssim,
            xshift=1.5cm,
    ylabel = Mean SSIM,
    ylabel near ticks,
    xlabel near ticks,
    title = SSIM on BAIR,
    legend columns=6,
    xtick distance=5,
    enlargelimits=false,
    y tick label style={
    /pgf/number format/.cd,
        fixed,
        fixed zerofill,
        precision=2,
    /tikz/.cd
    },
    width=\textwidth * 0.5,
    height=5cm,
    ymajorgrids
          ]

          \addplot [line width=0.5mm, color=glaucous, error bars/.cd, y dir=both, y explicit] table [x=t, y=oursmean, y error=ourserr, col sep=comma] {\bairpredtposessim};
  \addplot [line width=0.5mm, color=lava, error bars/.cd, y dir=both, y explicit] table [x=t, y=savpdetmean, y error=savpdeterr, col sep=comma]{\bairpredtposessim};
  \addplot [line width=0.5mm, color=black, error bars/.cd, y dir=both, y explicit] table [x=t, y=savpmean, y error=savperr, col sep=comma]{\bairpredtposessim};
  \addplot [line width=0.5mm, color=forestgreen, error bars/.cd, y dir=both, y explicit] table [x=t, y=svglpmean, y error=svglperr, col sep=comma]{\bairpredtposessim};
    \end{axis}
        \begin{axis}[    at={(axssim.south west)},
            name=axssim,
            yshift=-4.5cm,
    xlabel = Frame Number,
    ylabel = Mean PSNR,
    ylabel near ticks,
    xlabel near ticks,
    title = PSNR on BAIR,
    legend columns=6,
    xtick distance=5,
    enlargelimits=false,
    y tick label style={
    /pgf/number format/.cd,
        fixed,
        fixed zerofill,
        precision=2,
    /tikz/.cd
    },
    width=\textwidth * 0.5,
    height=5cm,
    ymajorgrids
          ]
          \addplot [line width=0.5mm, color=glaucous,  error bars/.cd, y dir=both, y explicit] table [x=t, y=oursmean, y error=ourserr, col sep=comma] {\bairpredtposepsnr};
  \addplot [line width=0.5mm, color=lava,  error bars/.cd, y dir=both, y explicit] table [x=t, y=savpdetmean, y error=savpdeterr, col sep=comma]{\bairpredtposepsnr};
    \addplot [line width=0.5mm, color=black, error bars/.cd, y dir=both, y explicit] table [x=t, y=savpmean, y error=savperr, col sep=comma]{\bairpredtposepsnr};
  \addplot [line width=0.5mm, color=forestgreen,  error bars/.cd, y dir=both, y explicit] table [x=t, y=svglpmean, y error=svglperr, col sep=comma]{\bairpredtposepsnr};
    \end{axis}        
  \end{tikzpicture}
  \caption{Per-frame visual metrics measured against video prediction propagation length. Lower is better for LPIPS scores, whereas higher is better for SSIM and PSNR. Our proposed method is generally competitive with other methods after 15 frames on LPIPS. }
  \label{fig:bair-photometric}
\end{figure}
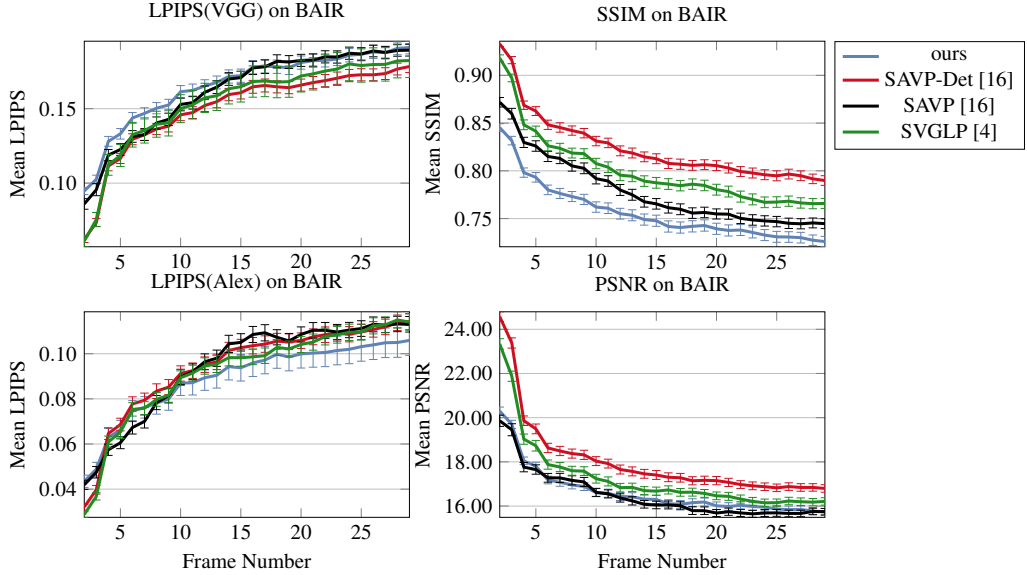

Our BAIR video prediction results are shown in Fig.~\ref{fig:bair-photometric}. Our LPIPS score (lower is better) starts higher than that of other methods, but become competitive after 15 frames. A likely explanation is because our method lacks skip connections from the input frame to the image decoder output, limiting our ability to perform pixel-copying behavior. Fig.~\ref{fig:bair-failure-analysis} shows output frames for all output models on a universally low-scoring sequence. Other methods have slight object blurring whereas ours remain sharp. However, one can also see that the objects in the back differ visually from the reference. Further, our red object disappears at frame 24. These, again are likely due to our information bottleneck, meaning predicting accurate dynamics in the pose representation is crucial for us. These fidelity issues are reflected in our consistently lower SSIM scores. 

\begin{figure}
  \centering
  \setlength\tabcolsep{0.32pt}
  \renewcommand{\arraystretch}{0.12}
  \begin{tabular}{cccccccccc}
    &t=1  & t=4& t=8& t=12 &t=16 &t=20& t=24& t=28& t=30\\\\
    \rotatebox{90}{~~~~~~GT}&
     \interpfig{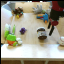}&
     \interpfig{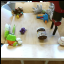}&
     \interpfig{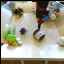}&
     \interpfig{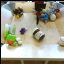}&
     \interpfig{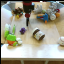}&
     \interpfig{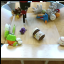}&
     \interpfig{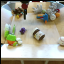}&
     \interpfig{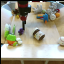}&
     \interpfig{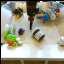}\\ 
     & \rotatebox{90}{\parbox{1.1cm}{Video \\ Flow}}&
     \interpfig{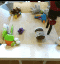}&
     \interpfig{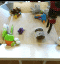}&
     \interpfig{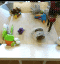}&
     \interpfig{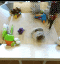}&
     \interpfig{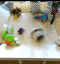}&
     \interpfig{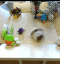}&
     \interpfig{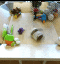}& \\
     & \rotatebox{90}{\parbox{1.1cm}{Super\\Slomo}}&
      \interpfig{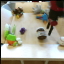}&
      \interpfig{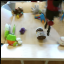}&
     \interpfig{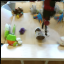}&
     \interpfig{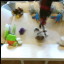}&
     \interpfig{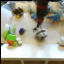}&
     \interpfig{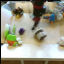}&
     \interpfig{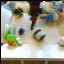}&\\
     & \rotatebox{90}{\parbox{1.1cm}{Ours}}&
      \interpfig{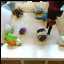}&
      \interpfig{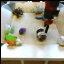}&
     \interpfig{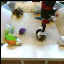}&
     \interpfig{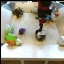}&
     \interpfig{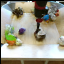}&
     \interpfig{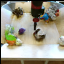}&
     \interpfig{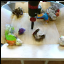}

  \end{tabular}
  \caption{\textbf{Latent-representation interpolation} on the test set. Linearly interpolating with our hidden pose representation follows a predictable path, similar to that which is produced by flow-based models such as ~\cite{jiang2018super}. ~\cite{kumar2019videoflow} is roughly able to interpolate, though it follows a less predictable path and may degrade in the middle. Frames for ~\cite{kumar2019videoflow} do not strictly correspond to the labeled time steps.}
  \label{Tbl:bair_interp_images}
\end{figure}

\boldhead{Video Interpolation}
We also experiment with 
long-range interpolation using our pose representation, interpolating between the first and last frame of test sequences. Qualitative results are shown in Fig. \ref{Tbl:bair_interp_images}. We compare with project page results from ~\cite{kumar2019videoflow} and a re-implementation of SuperSlomo~\cite{jiang2018super,AvinashSlomo}. While interpolation is not the primary goal of~\cite{kumar2019videoflow}, their latent-space interpolation is reasonably successful. However, as depicted in this example, their interpolated trajectory in image space is difficult to anticipate. Our pose-space interpolation results move predictably in a linear trajectory in image space while maintaining consistent image quality, closely aligning with the trajectory taken by the optical-flow-based SuperSlomo. Note that unlike our model and the VideoFlow model, the SuperSlomo model has never seen any training data from the BAIR push data, yet is able to reasonably translate the robot arm in the correct direction despite heavily degraded image quality and background distortion. We present a more detailed comparison against SuperSlomo in the appendix.

\subsection{KTH Action Dataset}
The KTH action dataset \cite{schuldt2004recognizing} consists of videos of people performing one of six actions (walking, running, jogging, boxing, handwaving, hand-clapping).
The background in these videos are fairly uniform, and human motion is regular.
However, this dataset presents challenges in walking, running and
jogging videos where the person comes in and out of the field of view. Further the camera zooms in and out in some action videos, creating non-static backgrounds. Consistent with \cite{lee2018stochastic}, we initialize our LSTM with 10 input frames and return the predicted sequence of the next 30. KTH
contains video sequences from 25 unique people with varying
appearance, split by person ids. 

\begin{figure}
  \begin{tikzpicture}[thick, scale=0.8]
    \begin{axis}[    name=ax1,
    ylabel = Mean LPIPS,
    ylabel near ticks,
    xlabel near ticks,
    title = LPIPS(VGG) on KTH,
    legend columns=6,
    xtick distance=5,
    enlargelimits=false,
    y tick label style={
    /pgf/number format/.cd,
        fixed,
        fixed zerofill,
        precision=2,
    /tikz/.cd
    },
    width=\textwidth * 0.5,
    height=5cm,
    ymajorgrids,
    legend columns=1,
    legend entries={ours, SAVP-Det \cite{lee2018stochastic}, SAVP \cite{lee2018stochastic}, DRNet \cite{denton2017unsupervised}},
    legend style={at={(2.6,1)},anchor=north}
      ]
  \addplot [line width=0.5mm, color=glaucous,error bars/.cd,y dir=both, y explicit] table [x=t, y=ours16_64vggmean, col sep=comma, y error=ours16_64vggstderr] {\kthpredtpose};
  \addplot [line width=0.5mm, color=lava,error bars/.cd,y dir=both, y explicit] table [x=t, y=savpdetvggmean, col sep=comma, y error=savpdetvggstderr]{\kthpredtpose};
  \addplot [line width=0.5mm, color=black,error bars/.cd,y dir=both, y explicit] table [x=t, y=savp0vggmean, col sep=comma,y error=savp0vggstderr]{\kthpredtpose};
    \addplot [line width=0.5mm, color=forestgreen,error bars/.cd,y dir=both, y explicit] table [x=t, y=drnetvggmean, col sep=comma,y error=drnetalexstderr]{\kthpredtpose};     
    \end{axis}
        \begin{axis}[    at={(ax1.south west)},
            yshift=-4.5cm,
    xlabel = Frame Number,
    ylabel = Mean LPIPS,
    ylabel near ticks,
    xlabel near ticks,
    title = LPIPS(Alex) on KTH,
    legend columns=6,
    xtick distance=5,
    enlargelimits=false,
    y tick label style={
    /pgf/number format/.cd,
        fixed,
        fixed zerofill,
        precision=2,
    /tikz/.cd
    },
    width=\textwidth * 0.5,
    height=5cm,
    ymajorgrids
          ]
          \addplot [line width=0.5mm, color=glaucous, error bars/.cd,y dir=both, y explicit] table [x=t, y=ours16_64alexmean, y error=ours16_64alexstderr,col sep=comma] {\kthpredtpose};
  \addplot [line width=0.5mm, color=lava, error bars/.cd,y dir=both, y explicit] table [x=t, y=savpdetalexmean, y error=savpdetalexstderr, col sep=comma]{\kthpredtpose};
  \addplot [line width=0.5mm, color=black,  error bars/.cd,y dir=both, y explicit] table [x=t, y=savp0alexmean, y error=savp0alexstderr, col sep=comma]{\kthpredtpose};
      \addplot [line width=0.5mm, color=forestgreen,error bars/.cd,y dir=both, y explicit] table [x=t, y=drnetalexmean, col sep=comma,y error=drnetalexstderr]{\kthpredtpose};     
    \end{axis}
        \begin{axis}[    at={(ax1.south east)},
            name=axssim,
            xshift=1.5cm,
    ylabel = Mean SSIM,
    ylabel near ticks,
    xlabel near ticks,
    title = SSIM on KTH,
    legend columns=6,
    xtick distance=5,
    enlargelimits=false,
    y tick label style={
    /pgf/number format/.cd,
        fixed,
        fixed zerofill,
        precision=2,
    /tikz/.cd
    },
    width=\textwidth * 0.5,
    height=5cm,
    ymajorgrids
          ]

          \addplot [line width=0.5mm, color=glaucous, error bars/.cd,y dir=both, y explicit] table [x=t, y=ours16_64ssim, y error=ours16_64ssimstderr,col sep=comma] {\kthpredtpose};
  \addplot [line width=0.5mm, color=lava ,error bars/.cd,y dir=both, y explicit] table [x=t, y=savpdetssim, y error=savpdetssimerr, col sep=comma]{\kthpredtpose};
  \addplot [line width=0.5mm, color=black,  error bars/.cd,y dir=both, y explicit] table [x=t, y=savpssim, y error=savpssimerr, col sep=comma]{\kthpredtpose};
   \addplot [line width=0.5mm, color=forestgreen,  error bars/.cd,y dir=both, y explicit] table [x=t, y=drnetssim, y error=drnetssimerr, col sep=comma]{\kthpredtpose};
    \end{axis}
        \begin{axis}[    at={(axssim.south west)},
            name=axssim,
            yshift=-4.5cm,
    xlabel = Frame Number,
    ylabel = Mean PSNR,
    ylabel near ticks,
    xlabel near ticks,
    title = PSNR on KTH,
    legend columns=6,
    xtick distance=5,
    enlargelimits=false,
    y tick label style={
    /pgf/number format/.cd,
        fixed,
        fixed zerofill,
        precision=2,
    /tikz/.cd
    },
    width=\textwidth * 0.5,
    height=5cm,
    ymajorgrids
          ]
          \addplot [line width=0.5mm, color=glaucous, error bars/.cd,y dir=both, y explicit] table [x=t, y=ours16_64psnr, y error=ours16_64psnrstderr,col sep=comma] {\kthpredtpose};
  \addplot [line width=0.5mm, color=lava ,error bars/.cd,y dir=both, y explicit] table [x=t, y=savpdetpsnr, y error=savpdetpsnrerr, col sep=comma]{\kthpredtpose};
  \addplot [line width=0.5mm, color=black,  error bars/.cd,y dir=both, y explicit] table [x=t, y=savppsnr, y error=savppsnrerr, col sep=comma]{\kthpredtpose};
   \addplot [line width=0.5mm, color=forestgreen,  error bars/.cd,y dir=both, y explicit] table [x=t, y=drnetpsnr, y error=drnetpsnrerr, col sep=comma]{\kthpredtpose};
    \end{axis}
  \end{tikzpicture}
  \caption{Per-frame visual metrics measured against video propagation length on the KTH dataset. Results are evaluated on image frames downsampled to $64\times64$. DRNet results were downsampled from $128 \times 128$ to match the target resolution.}
  \label{fig:kth-photometric}
\end{figure}
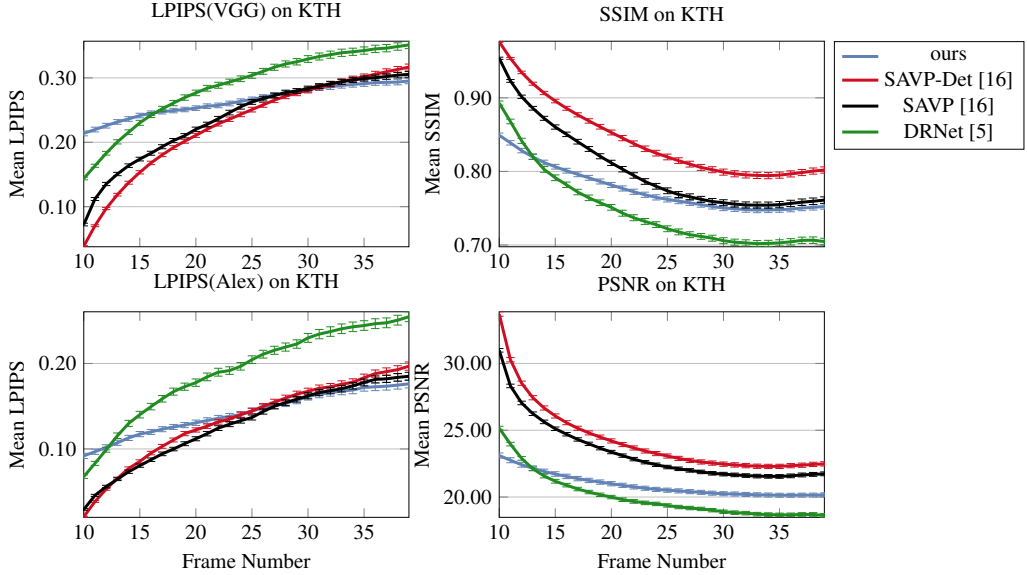

\begin{figure}
  
  \setlength\tabcolsep{0.32pt}
  \renewcommand{\arraystretch}{0.15}
  \centering
  \begin{tabular}{rccc||ccc||ccc}
    &t=11  &t=23 &t=35 &t=11  &t=23 &t=35 &t=11&t=23&t=35\\\\
         \rotatebox{90}{GT}&
         \interpfigkth{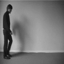}&
     \interpfigkth{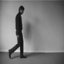}&
     \interpfigkth{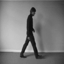}&
     \interpfigkth{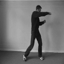}&
     \interpfigkth{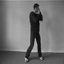}&
     \interpfigkth{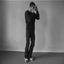}&
     \interpfigkth{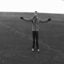}&
     \interpfigkth{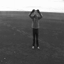}&
     \interpfigkth{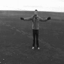}\\
     \rotatebox{90}{Ours}&
     \interpfigkth{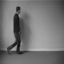}&
     \interpfigkth{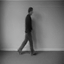}&
     \interpfigkth{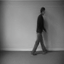}&
     \interpfigkth{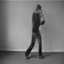}&
     \interpfigkth{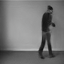}&
     \interpfigkth{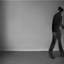}&
     \interpfigkth{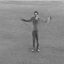}&
     \interpfigkth{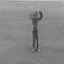}&
     \interpfigkth{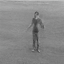}\\
     \rotatebox{90}{\parbox{0.9cm}{~Savp \\~Det}}&
          \interpfigkth{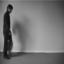}&
     \interpfigkth{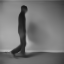}&
     \interpfigkth{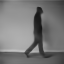}&
     \interpfigkth{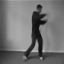}&
     \interpfigkth{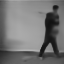}&
     \interpfigkth{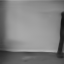}&
     \interpfigkth{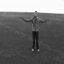}&
     \interpfigkth{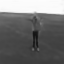}&
     \interpfigkth{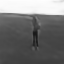}
     \\
     \rotatebox{90}{~Savp}&
          \interpfigkth{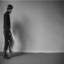}&
     \interpfigkth{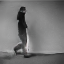}&
     \interpfigkth{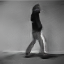}&
     \interpfigkth{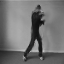}&
     \interpfigkth{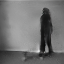}&
     \interpfigkth{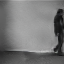}&
     \interpfigkth{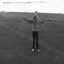}&
     \interpfigkth{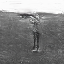}&
     \interpfigkth{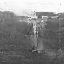}\\
     \rotatebox{90}{DRNet}&
     \interpfigkth{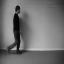}&
     \interpfigkth{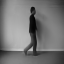}&
     \interpfigkth{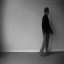}&
     \interpfigkth{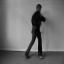}&
     \interpfigkth{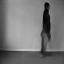}&
     \interpfigkth{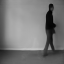}&
     \interpfigkth{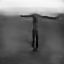}&
     \interpfigkth{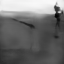}&
     \interpfigkth{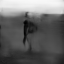}\\

  \end{tabular}
  \caption{Worst sequence analysis. From left to right, our best sequence, one of the worst sequences (based on LPIPS scores) shared by both SAVP\cite{lee2018stochastic} and SAVP Det, and one of our worst performing sequences. Our proposed method may not reproduce the foreground and background aas accurately, but appears to maintain better structural integrity.}
  \label{fig:kth_vpred}
\end{figure}

Quantitative video prediction results are shown in Fig.~\ref{fig:kth-photometric}. Consistent with results on BAIR, our method appears to be competitive after roughly 15 frames of prediction. From left to right, the qualitative results shown in Fig.~\ref{fig:kth_vpred} depict one of our strong examples, a weak example for SAVP\cite{lee2018stochastic} and SAVP-Det, and one of the lowest scoring examples for our method. Compared to SAVP, both stochastic and deterministc, our model is better at maintaining structural integrity in these cases. In the last column, notice our model completely fails to reproduce the correct background. SAVP and SAVP-Det are able to faithfully reproduce the skyline, but their foreground object loses structure nearing the end. While DRNet~\cite{denton2017unsupervised}, like our work, focuses on disentangling pose and appearance information, they struggle more with maintaining structural integrity. On the other hand, they do a better job of preserving the skyline in the rightmost failure case than our method.

%% file: 4-sec_related_work.tex
\section{Related Work}
Several deep learning methods have been proposed for future frame prediction and video frame interpolation.
One common approach for both of these tasks is to model motion transformation via flow.
Finn et al.~\cite{finn2016unsupervised} proposed an LSTM to model motion transformation over via kernel-based transformation directly on pixels. 
Other works extrapolate flow-based motion into the future frames \cite{xue2016visual, walker2015dense,reda2018sdc,vondrick2017generating} or use it to interpolate it between frames \cite{jiang2018super, liu2017video, reda2019unsupervised}.
Another line of research seeks to improve video prediction via different loss functions. Mathieu et al.~\cite{ mathieu2015deep} shows that training a future frame generator with adversarial loss can produce sharper images than L2-based losses. The use of stochastic models to learn the uncertainty in the motion to effectively reason over uncertain futures has also garnered recent attention.
Among these, variational auto encoders (VAEs) \cite{kingma2013auto} have been widely explored  \cite{babaeizadeh2017stochastic, denton2018stochastic, lee2018stochastic}.
With the recent success of the Flow based models \cite{dinh2016density, kingma2018glow} in image generation, \cite{pottorff2019video, kumar2019videoflow} propose to optimize the likelihood of training videos.
In this work, we model the future pose prediction in a deterministic way, however, it can be extended to a stochastic model with VAEs or Flow based models for temporal prediction. 

Several related works explore the video data manipulation and extrapolation in factored pose and appearance representations~\cite{wichers2018hierarchical,tulyakov2018mocogan,denton2017unsupervised,siarohin2018animating}. DRNet~\cite{denton2017unsupervised} and Wichers et al.\cite{wichers2018hierarchical}, which share our general approach to factored representations, use temporal prediction to propagate pose into the future, and image reconstruction from new pose information while \cite{denton2017unsupervised} uses a set of novel adversarial losses to separate appearance from pose. While these learned encodings exhibit properties of appearance and pose, the representation is high-dimensional and unconstrained with dynamics that can be difficult to learn and therefore prone to degradation over long-range predictions. Wichers et al.~\cite{wichers2018hierarchical} learn an implicit, yet high dimensional, representation with coordinate-like properties \cite{reed2015deep}. \cite{siarohin2018animating} similarly uses unsupervised landmarks as a latent representation for video content manipulation, but uses an image-warping paradigm instead of directly rendering new frames from the compact latent representation.

Works under unsupervised discovery of image correspondences\cite{thewlis2017unsupervised,Zhang_2018_CVPR,suwajanakorn2018discovery,Thewlis17,Kanazawa_2016_CVPR,jakab2018unsupervised,lorenz2019unsupervised} form the backbone of this work. Most relevant here are~\cite{jakab2018unsupervised} and \cite{lorenz2019unsupervised}, which learn the latent landmark activation maps via an image factorization and reconstruction pipeline. Each image is factored into pose and appearance representations and a decoder is trained to reconstruct the image from these latent factors. The loss is designed such that accurate image reconstruction can only be achieved when the landmarks activate at consistent locations between an image its TPS-warped variant. Lorenz et al.\cite{lorenz2019unsupervised} specifically improves upon the method proposed by \cite{jakab2018unsupervised} such that instead of representing the appearance information as a single vector for the entire image, there is a separate appearance encoding for each landmark in the pose representation. Our work in turn builds on \cite{lorenz2019unsupervised} by demonstrating how to safely manipulate the 2D Gaussian-based pose representation in a learned temporal dynamics module, as well as improving their landmark quality by incorporating a decoder specialized for conditioning on semantic maps~\cite{park2019semantic}.

%% file: 5-sec_conclusion.tex
\section{Discussion and Conclusion}
We present a fully unsupervised paradigm for long range video
prediction and interpolation that is capable of preserving image
structure over long time frames. We achieve this by factorizing image
data into pose information and temporally invariant appearance
information. Crucially, as the pose information will go through
numerous perturbations throughout long-range video prediction, we
use of collections of unsupervised 2D Gaussian landmarks
as our pose state. We demonstrate that we can safely perturb each 2D Gaussian in
Gaussian parameter space ($\mu$, $\Sigma$) before projecting to 2D
heatmaps to be fed into the image decoder, thereby guaranteeing that the
image decoder will always see a collection of unimodal Gaussian heatmaps.

\boldhead{Limitations and Future Work} The pose and appearance factorization creates an
  information bottleneck within the pipeline, limiting our upper bound
  on pixel-wise fidelity. As previously noted, the foreground object's appearance often varies drastically from the reference sequence as our pipeline has difficulty rendering novel appearances. This is less noticable on datasets such as BAIR where the appearances vary little between training and test data, but is easily noticable on the BBC and KTH results (see supplemental material). Further, because the landmark learning process encourages
  assigning landmarks tend to favor moving foreground patterns, the background is largely unhandled by our model.

  Our temporal prediction is stable for a little over 100 frames, after which it starts to degrade. The temporal model could predict a valid pose state that the LSTM simply has not seen before, leading unexpected behavior. Nevertheless, we believe that a landmark-based approach will open doors to longer, more realistic predictions, as it is possible to explicitly model the trajectories of landmarks in an explicit physics simulator -- something which is nearly impossible using less interpretable representations.

\boldhead{Acknowledgements} We would like to thank Alex Lee and Emily Denton for releasing the source code and pre-trained weights for their respective models. We would also like to thank Dominik Lorenz for assisting us in reproducing their method via email correspondence.

%% file: 6-appendix.tex
\appendix
\section{Implementation Details}
The overall architecture consists of 3 sub-networks as in \cite{lorenz2019unsupervised}: pose and appearance encoding networks and a generator network for image reconstruction. 
We use the U-net architectures \cite{ronneberger2015u} for the pose and appearance encoders.
The pose encoder has 4 blocks of convolutional dowsampling modules.
Each convolutional downsampling module has a convolution layer-Instance Normalization-ReLU and a downsampling layer. At each block, the number of filters doubles, starting from 64. The upsampling portion of the pose encoder has 3 blocks of convolutional upsampling modules, and the number of channels is halved at every block starting from 512.
The appearance encoder network has one convolutional downsampling module and one convolutional upsampling module. Bothe pose and appearance encoder networks feature skip layers.

The image decoder has 4 convolution-ReLU-upsample modules. We first downsample the appearance featuremap by a factor of 8 in each spatial dimension. Number of output channels of each convolution-ReLU-upsampling module in the image decoder are 256, 256, 128, 64, and 3 respectively. We apply spectral normalization \cite{miyato2018spectral} to each convolutional layer.

Our LSTM comprises 3 LSTM layers and a final linear layer. Each LSTM layer has 256 channels.
For brevity, we specify our LSTM training setups
in the format of $n$ inputs $m$ future. This means the LSTM is
required to predict a supervised sequence of $n+m-1$ frames, using
ground truth input for the first $n$ timesteps and its own previous
timestep's output for the remaining $m-1$.

We set our models to learn 40 latent landmarks for BBC, and 30 for BAIR and KTH.

\boldhead{BBC} Prior to training our model, we first roughly crop around each signer
and resize to 128$\times$128. For video prediction training, we use a
10 input 10 future setup.

\boldhead{BAIR} Our video prediction LSTM is trained with a 10 input 0 future setup (never
conditions on its own output during training).

\boldhead{KTH} Our video prediction LSTM is trained with a 10 input 10 future setup.

\boldhead{VGG Perceptual Loss} Our VGG Perceptual loss uses the pre-trained VGG19 model provided by the PyTorch library. We apply the MSE loss on outputs of layers \texttt{relu1\_2}, \texttt{relu2\_2}, \texttt{relu3\_2}, and \texttt{relu4\_2}, weighted by $\frac{1}{32}$,$\frac{1}{16}$,$\frac{1}{8}$,and $\frac{1}{4}$ respectively.

\boldhead{Adversarial Loss} The image decoder $\mathit{Dec}$ receives additional supervision from a fully convolutional discriminator $D$. $D$ aims to distinguish real images from the reconstructed ones. The objective of adversarial loss is given by
$\log{D(x)+\log(1-D(\mathit{Dec}(\tilde{\Phi}^{pose},\Phi^{app})))}$ where $x$ is a real image.
We use a 3 layer discriminator. Each layer has a convolutional layer of 64, 128, and 256 channels respecively, an instance normalization layer, and a leaky relu.

\boldhead{Optimization Parameters} We train the image decoder with Adam optimizer, learning rate of $1e^{-4}$. We use weight decay of $5e^{-6}$. Each image decoder is trained on 8 GPUs with batch size of 16 images per GPU.

\newpage
\section{BBC Pose Key Point Evaluation}
\boldhead{Linear Regression test Implementation Details}
We find the center of the tightest bounding box
surrounding the annotated keypoints and dilate it by 150 pixels in each
direction before resizing to 128$\times$128 to feed into our network. Next, we translate the coordinate spaces for both our predicted landmark locations and the target ground truth keypoints
such that the origin is always centered within the crop, then fit a
linear regressor without intercept to learn the mapping.

\begin{figure}[h]
\includegraphics[width=\textwidth, trim={3cm, 0, 3cm, 0}, clip]{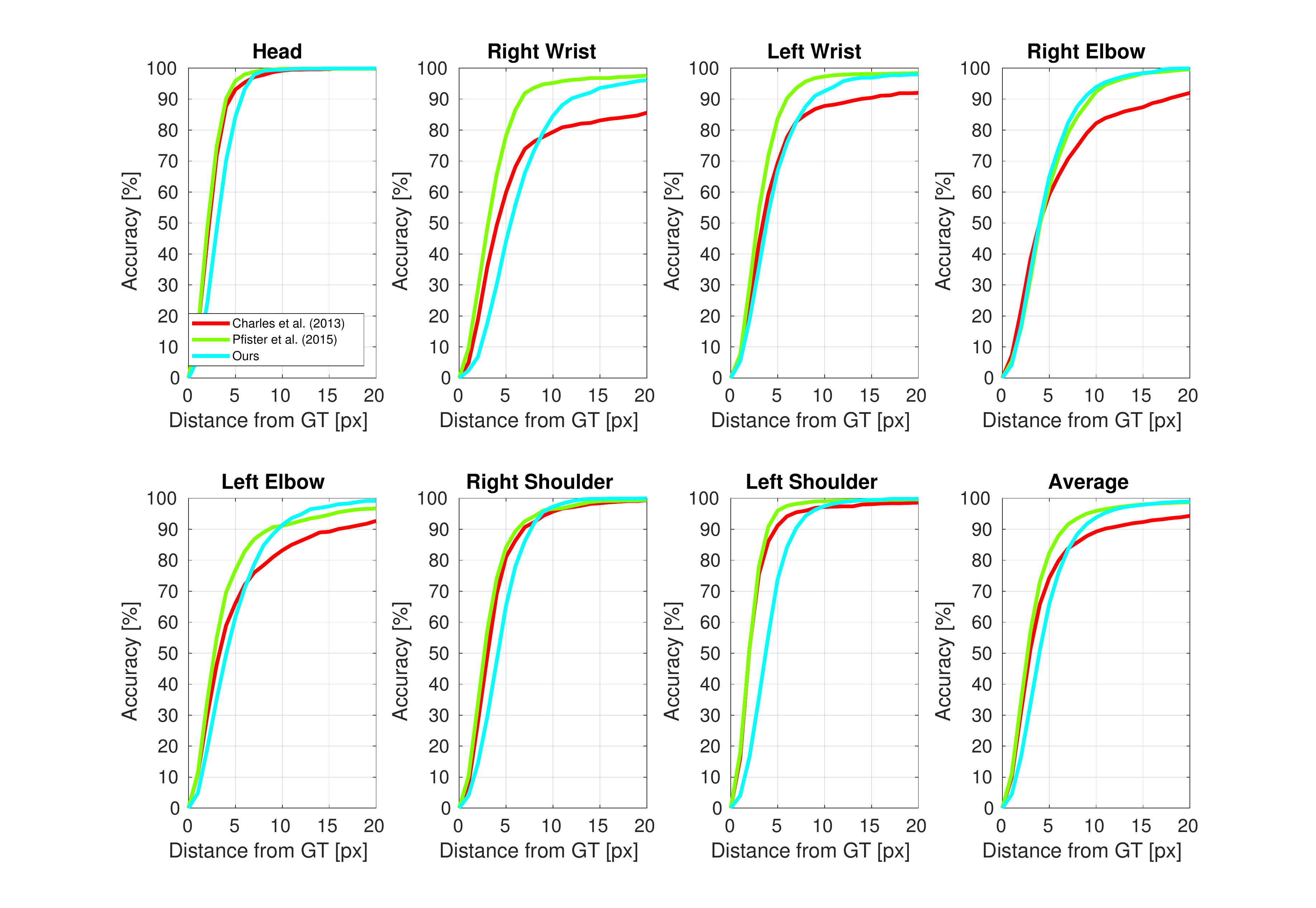}
\caption{Accuracy vs distance in pixels on the test set.}
\end{figure}

\newpage
\section{BBC Pose Video Prediction Results}
Here we show additional qualitative results on the BBC dataset. These are easier predictions because the initialization frames are held-out frames from the training set. This means that the model does not see these exact frames during training, but has access to visually similar frames from the same sequence.

\begin{figure}[h]
  \centering
  \setlength\tabcolsep{0.32pt}
  \renewcommand{\arraystretch}{0.15}
  \begin{tabular}{cccccccccc}
    &t=10 &t=20 &t=30 &t=40 &t=50 &t=60 &t=70 &t=80 &t=90\\\\
      \rotatebox{90}{~~~GT}&
     \interpfig{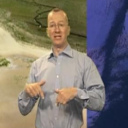}&
     \interpfig{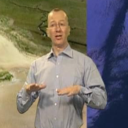}&
     \interpfig{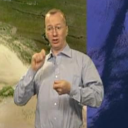}&
     \interpfig{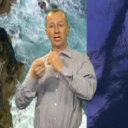}&
     \interpfig{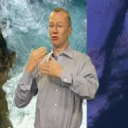}&
     \interpfig{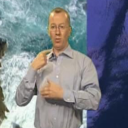}&
    \interpfig{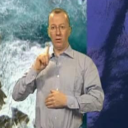}&
    \interpfig{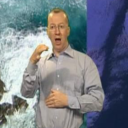}&
     \interpfig{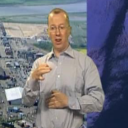}\\
    \rotatebox{90}{~~~Ours}&
     \interpfig{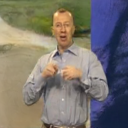}&
     \interpfig{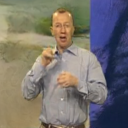}&
     \interpfig{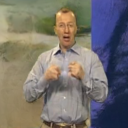}&
     \interpfig{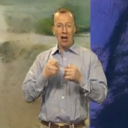}&
     \interpfig{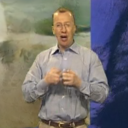}&
     \interpfig{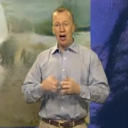}&
    \interpfig{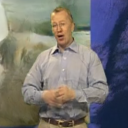}&
    \interpfig{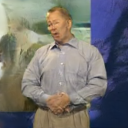}&
     \interpfig{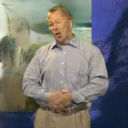}\\
    \rotatebox{90}{~~~GT}&
     \interpfig{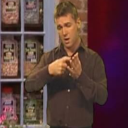}&
     \interpfig{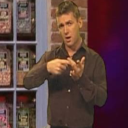}&
     \interpfig{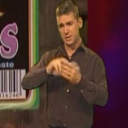}&
     \interpfig{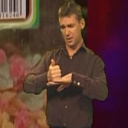}&
     \interpfig{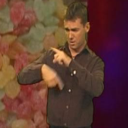}&
     \interpfig{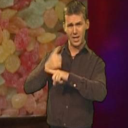}&
     \interpfig{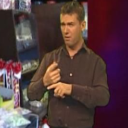}&
     \interpfig{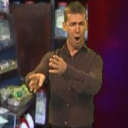}&
     \interpfig{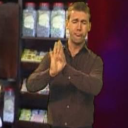}\\
    \rotatebox{90}{~~~Ours}&
     \interpfig{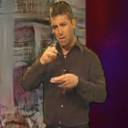}&
     \interpfig{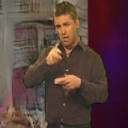}&
     \interpfig{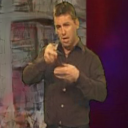}&
     \interpfig{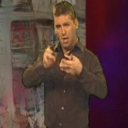}&
     \interpfig{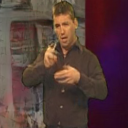}&
     \interpfig{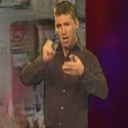}&
    \interpfig{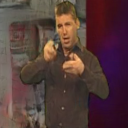}&
    \interpfig{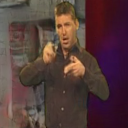}&
    \interpfig{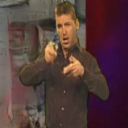}\\
        \end{tabular}
     \caption{Additional BBC video prediction results for long range video prediction. We display predictions of every 10th frame from held out examples of BBC-Pose training set. The LSTM is conditioned on the first 2 video frames and used to predict the next 98}
  \label{fig:bbc_pred_app}
\end{figure}

\newpage
\section{Failure cases: BBC Pose Video Prediction Results with Novel Appearances}
This section features a more difficult task of predicting video sequences for novel appearances on the validation set. This means that the signer has been seen by the model in the training data, but the attire is novel. Unfortunately the rendering pipeline is not sufficiently generalized for this task, having only trained on the appearances of 5 signer-attire pairings. As such, the predicted foreground varies drastically from the reference -- more closely resembling a signer-attire pairing from the training set. We are interested in addressing this limitation in future work.

\begin{figure}[h]
  \centering
  \setlength\tabcolsep{0.32pt}
  \renewcommand{\arraystretch}{0.15}
  \begin{tabular}{cccccccccc}
    &t=10 &t=20 &t=30 &t=40 &t=50 &t=60 &t=70 &t=80 &t=90\\\\
    
    \rotatebox{90}{~~~GT}&
     \interpfig{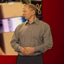}&
     \interpfig{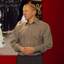}&
     \interpfig{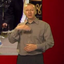}&
     \interpfig{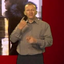}&
     \interpfig{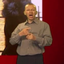}&
     \interpfig{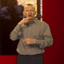}&
    \interpfig{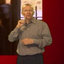}&
     \interpfig{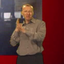}&
    \interpfig{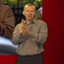}\\
    \rotatebox{90}{~~~Ours}&
     \interpfig{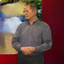}&
     \interpfig{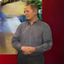}&
     \interpfig{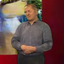}&
     \interpfig{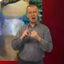}&
     \interpfig{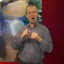}&
     \interpfig{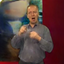}&
    \interpfig{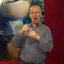}&
     \interpfig{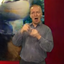}&
    \interpfig{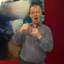}\\
        \rotatebox{90}{~~~GT}&
     \interpfig{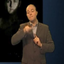}&
     \interpfig{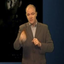}&
     \interpfig{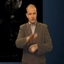}&
     \interpfig{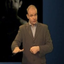}&
     \interpfig{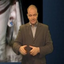}&
     \interpfig{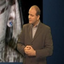}&
    \interpfig{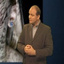}&
     \interpfig{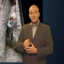}&
    \interpfig{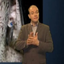}\\
    \rotatebox{90}{~~~Ours}&
     \interpfig{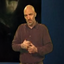}&
     \interpfig{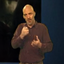}&
     \interpfig{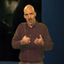}&
     \interpfig{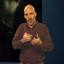}&
     \interpfig{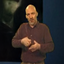}&
     \interpfig{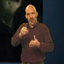}&
    \interpfig{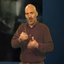}&
     \interpfig{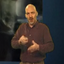}&
    \interpfig{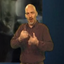}\\
    \rotatebox{90}{~~~GT}&
     \interpfig{Video_pred/BBC_val_vid13/GT/000010.png}&
     \interpfig{Video_pred/BBC_val_vid13/GT/000020.png}&
     \interpfig{Video_pred/BBC_val_vid13/GT/000030.png}&
     \interpfig{Video_pred/BBC_val_vid13/GT/000040.png}&
     \interpfig{Video_pred/BBC_val_vid13/GT/000050.png}&
     \interpfig{Video_pred/BBC_val_vid13/GT/000060.png}&
    \interpfig{Video_pred/BBC_val_vid13/GT/000070.png}&
     \interpfig{Video_pred/BBC_val_vid13/GT/000080.png}&
    \interpfig{Video_pred/BBC_val_vid13/GT/000090.png}\\
    \rotatebox{90}{~~~Ours}&
     \interpfig{Video_pred/BBC_val_vid13/Ours/000010.png}&
     \interpfig{Video_pred/BBC_val_vid13/Ours/000020.png}&
     \interpfig{Video_pred/BBC_val_vid13/Ours/000030.png}&
     \interpfig{Video_pred/BBC_val_vid13/Ours/000040.png}&
     \interpfig{Video_pred/BBC_val_vid13/Ours/000050.png}&
     \interpfig{Video_pred/BBC_val_vid13/Ours/000060.png}&
    \interpfig{Video_pred/BBC_val_vid13/Ours/000070.png}&
     \interpfig{Video_pred/BBC_val_vid13/Ours/000080.png}&
    \interpfig{Video_pred/BBC_val_vid13/Ours/000090.png}\\
    \rotatebox{90}{~~~GT}&
     \interpfig{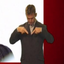}&
     \interpfig{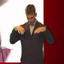}&
     \interpfig{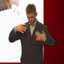}&
     \interpfig{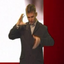}&
     \interpfig{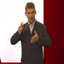}&
     \interpfig{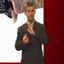}&
    \interpfig{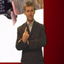}&
     \interpfig{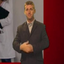}&
    \interpfig{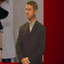}\\
    \rotatebox{90}{~~~Ours}&
     \interpfig{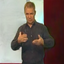}&
     \interpfig{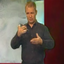}&
     \interpfig{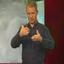}&
     \interpfig{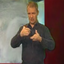}&
     \interpfig{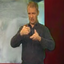}&
     \interpfig{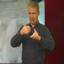}&
    \interpfig{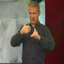}&
     \interpfig{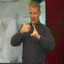}&
    \interpfig{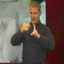}\\
   \rotatebox{90}{~~~GT}&
     \interpfig{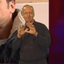}&
     \interpfig{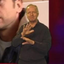}&
     \interpfig{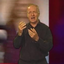}&
     \interpfig{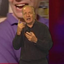}&
     \interpfig{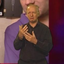}&
     \interpfig{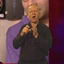}&
    \interpfig{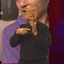}&
     \interpfig{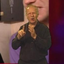}&
    \interpfig{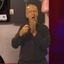}\\
    \rotatebox{90}{~~~Ours}&
     \interpfig{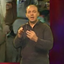}&
     \interpfig{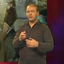}&
     \interpfig{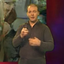}&
     \interpfig{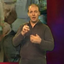}&
     \interpfig{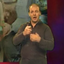}&
     \interpfig{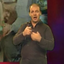}&
    \interpfig{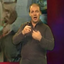}&
     \interpfig{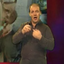}&
    \interpfig{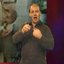}\\
    \end{tabular}
     \caption{Qualitative results for long range video prediction. The appearance frames come from the validation set where the signers appear with different clothing. In our experiments on this dataset, pose encoding generalizes well as it has seen a variety of poses. On the other hand, the appearance encoder does not generalize well because of the limited number of appearances it has seen in the training set. The output frames while being realistic, do not always match the reference appearance.}
  \label{fig:bbc_pred_app}
\end{figure}

\newpage
\section{BAIR Dataset - Video Interpolation}

\begin{figure}[h]
  \centering
  \setlength\tabcolsep{0.32pt}
  \renewcommand{\arraystretch}{0.12}
  \begin{tabular}{cccccccccc}
    &t=1  & t=4& t=8& t=12 &t=16 &t=20& t=24& t=28& t=30\\\\
    \rotatebox{90}{~~~~~~GT}&
     \interpfig{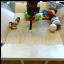}&
     \interpfig{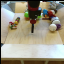}&
     \interpfig{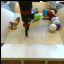}&
     \interpfig{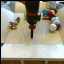}&
     \interpfig{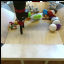}&
     \interpfig{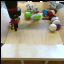}&
     \interpfig{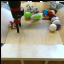}&
     \interpfig{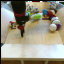}&
     \interpfig{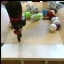}\\ 
     & \rotatebox{90}{\parbox{1.5cm}{\vspace{0.9cm}~~~Video \\~~~~Flow\cite{kumar2019videoflow}}}&
     \interpfig{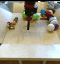}&
     \interpfig{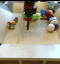}&
     \interpfig{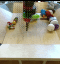}&
     \interpfig{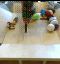}&
     \interpfig{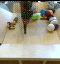}&
     \interpfig{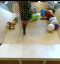}&
     \interpfig{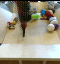}& \\
     & \rotatebox{90}{\parbox{1.5cm}{\vspace{0.9cm}~~Super\\Slomo\cite{jiang2018super}}}&
      \interpfig{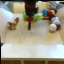}&
      \interpfig{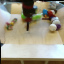}&
     \interpfig{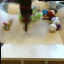}&
     \interpfig{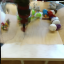}&
     \interpfig{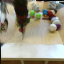}&
     \interpfig{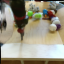}&
     \interpfig{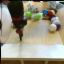}&\\
     & \rotatebox{90}{\parbox{1.5cm}{\vspace{0.9cm}~~~~Ours}}&
      \interpfig{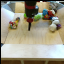}&
      \interpfig{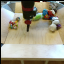}&
     \interpfig{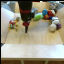}&
     \interpfig{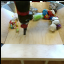}&
     \interpfig{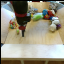}&
     \interpfig{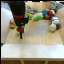}&
     \interpfig{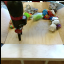}

  \end{tabular}
  \caption{\textbf{Latent-representation interpolation} Additional qualitative interpolation results on the test set.}
  \label{fig:bair_interp_app}
\end{figure}

\pgfplotstableread[col sep=comma]{data/bair_interp.txt}\bairinterpdata
\pgfplotstabletranspose[input colnames to=t,colnames from=t]\bairinterpdatatpose\bairinterpdata

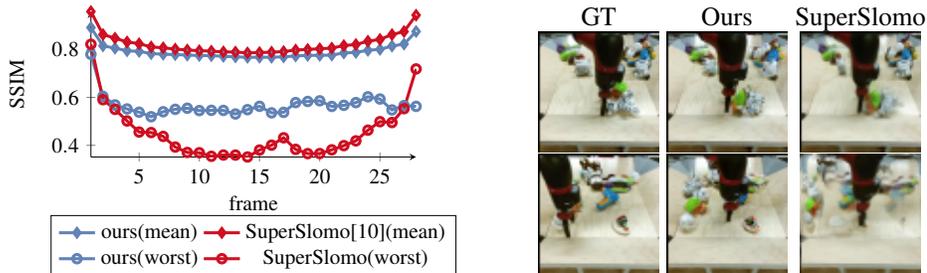
\begin{figure}[h]
\begin{minipage}{0.50\textwidth}
\begin{tikzpicture}[thick, scale=0.8]
\begin{axis}[
    axis lines = left,
    xlabel = frame,
    ylabel = SSIM,
    legend columns=2,
    width=\textwidth,
    height=4cm,
    legend style={at={(0.5,-0.4)},anchor=north}
]
  \addplot [line width=0.5mm, color=glaucous, mark=diamond] table [x=t, y=oursmean, col sep=comma] {\bairinterpdatatpose};
  \addplot [line width=0.5mm, color=lava, mark=diamond] table [x=t, y=SuperSlomomean, col sep=comma]{\bairinterpdatatpose};
  \addplot [line width=0.5mm, color=glaucous, mark=o] table [x=t, y=oursworst, col sep=comma]{\bairinterpdatatpose};
  \addplot [line width=0.5mm, color=lava, mark=o]table [x=t, y=SuperSlomoworst, col sep=comma] {\bairinterpdatatpose};
  \legend{ours(mean), SuperSlomo\cite{jiang2018super}(mean), ours(worst), SuperSlomo(worst)}
\end{axis}
\end{tikzpicture}
\end{minipage}
\begin{minipage}{0.4\textwidth}
  \setlength\tabcolsep{1.5pt}
  \renewcommand{\arraystretch}{0.24}
  \begin{tabular}{ccc}
    GT & Ours & SuperSlomo\\
    \interpfiglarge{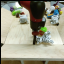}&
    \interpfiglarge{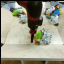}&
    \interpfiglarge{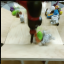}\\
    \interpfiglarge{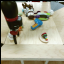}&
    \interpfiglarge{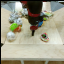}&
    \interpfiglarge{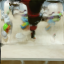}
  \end{tabular}
\end{minipage}

\caption{Left: Long-range interpolation SSIM on BAIR. We interpolate our
  hidden pose representations between the first and last frame, and
  evaluate the SSIM of the decoded intermediate frames. We compare
  against an off-the-shelf SuperSlomo implementation. Right: Each row compares the the lowest SSIM frames of either our model (top) or SuperSlomo (bottom). Notice how even when SSIM is low, our image is still visually plausible. Our loss in SSIM is largely due to poor pixel fidelity due to our lack of a skip-layer-like mechanisem from the input frames. SuperSlomo has high pixel fidelity, as it operates directly on the input pixels and can achieve near-perfect background when motion is minimal. However, its failure cases involve casastrophic warping of the image (see background distortion in bottom row example).}
  \label{fig:ss_interp}
\end{figure}

\begin{figure}[h]
  \centering
  \setlength\tabcolsep{0.32pt}
  \renewcommand{\arraystretch}{0.12}
  \begin{tabular}{cccccccccc}
    &t=1  & t=4& t=8& t=12 &t=16 &t=20& t=24& t=28& t=30\\\\
    \rotatebox{90}{~~~~~~GT}&
     \interpfig{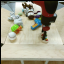}&
     \interpfig{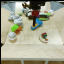}&
     \interpfig{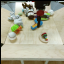}&
     \interpfig{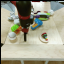}&
     \interpfig{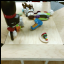}&
     \interpfig{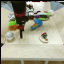}&
     \interpfig{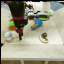}&
     \interpfig{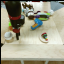}&
     \interpfig{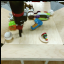}\\ 
     & \rotatebox{90}{\parbox{1.5cm}{\vspace{0.9cm}~~Super\\Slomo\cite{jiang2018super}}}&
      \interpfig{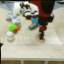}&
      \interpfig{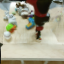}&
     \interpfig{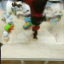}&
     \interpfig{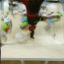}&
     \interpfig{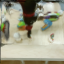}&
     \interpfig{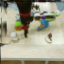}&
     \interpfig{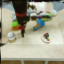}&\\
     & \rotatebox{90}{\parbox{1.5cm}{\vspace{0.9cm}~~~~Ours}}&
      \interpfig{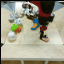}&
      \interpfig{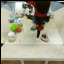}&
     \interpfig{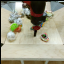}&
     \interpfig{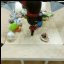}&
     \interpfig{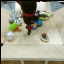}&
     \interpfig{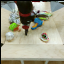}&
     \interpfig{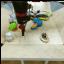}\\
         &t=1  & t=4& t=8& t=12 &t=16 &t=20& t=24& t=28& t=30\\\\
    \rotatebox{90}{~~~~~~GT}&
     \interpfig{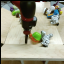}&
     \interpfig{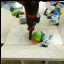}&
     \interpfig{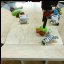}&
     \interpfig{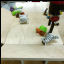}&
     \interpfig{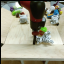}&
     \interpfig{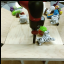}&
     \interpfig{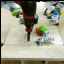}&
     \interpfig{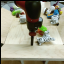}&
     \interpfig{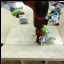}\\ 
     & \rotatebox{90}{\parbox{1.5cm}{\vspace{0.9cm}~~Super\\Slomo\cite{jiang2018super}}}&
      \interpfig{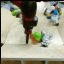}&
      \interpfig{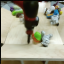}&
     \interpfig{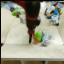}&
     \interpfig{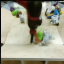}&
     \interpfig{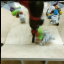}&
     \interpfig{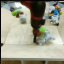}&
     \interpfig{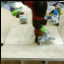}&\\
     & \rotatebox{90}{\parbox{1.5cm}{\vspace{0.9cm}~~~~Ours}}&
      \interpfig{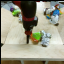}&
      \interpfig{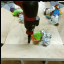}&
     \interpfig{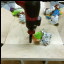}&
     \interpfig{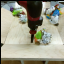}&
     \interpfig{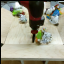}&
     \interpfig{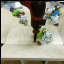}&
     \interpfig{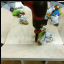}

  \end{tabular}
  \caption{\textbf{Latent-representation interpolation} on the test set. The top video is from the test sequences with the worst SSIM from SuperSlomo ~\cite{jiang2018super}. The bottom video sequence is the one with the worst SSIM from our model.}
  \label{fig:bair_interp_app_ss_comp1}
\end{figure}

In Fig. \ref{fig:ss_interp}, we report SSIM results over the BAIR test set in interpolation task.
Our method achieves comparable accuracy with SuperSlomo while being slightly worse in the first and last few frames.
One main difference between these two methods is that SuperSlomo is able to copy over the last frame, and perfectly translate the stationary pixels whereas our pose and appearance factorization creates an information bottleneck and hardens the perfect image reconstruction.
The SSIM metric is more sensitive to the differences in reconsruction errors than human perception.
As such, in Fig.  \ref{fig:ss_interp}, we share the frames that has the worst SSIM score for our method in the top and SuperSlomo in the second row.
While the generated frame by our method with the lowest SSIM score is still visually plausible, the one from SuperSlomo has blurred background and removed objects. We show more frames from this sequence for these two videos in Fig. \ref{fig:bair_interp_app_ss_comp1}.

\clearpage

\newpage
\section{BAIR Dataset - Additional Video Prediction Results}

\begin{figure}[h]
  \centering
  \setlength\tabcolsep{0.32pt}
  \renewcommand{\arraystretch}{0.12}
  \begin{tabular}{cccccccccc}
    &t=2  & t=4& t=8& t=12 &t=16 &t=20& t=24& t=28& t=30\\
     \rotatebox{90}{~~~GT}&
     \interpfig{Video_pred/BAIR_iter116/GT/2.png}&
     \interpfig{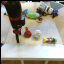}&
     \interpfig{Video_pred/BAIR_iter116/GT/7.png}&
     \interpfig{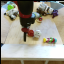}&
     \interpfig{Video_pred/BAIR_iter116/GT/15.png}&
     \interpfig{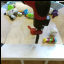}&
     \interpfig{Video_pred/BAIR_iter116/GT/23.png}&
     \interpfig{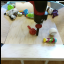}&
     \interpfig{Video_pred/BAIR_iter116/GT/29.png}\\
          & Ours &
      \interpfig{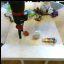}&
      \interpfig{Video_pred/BAIR_iter116/Ours/000007.png}&
     \interpfig{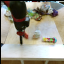}&
     \interpfig{Video_pred/BAIR_iter116/Ours/000015.png}&
     \interpfig{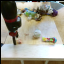}&
     \interpfig{Video_pred/BAIR_iter116/Ours/000023.png}&
     \interpfig{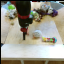}&
     \interpfig{Video_pred/BAIR_iter116/Ours/000029.png}\\
       & SVGLP &
     \interpfig{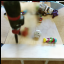}&
     \interpfig{Video_pred/BAIR_iter116/SVGLP/7.png}&
     \interpfig{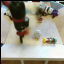}&
     \interpfig{Video_pred/BAIR_iter116/SVGLP/15.png}&
     \interpfig{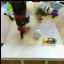}&
     \interpfig{Video_pred/BAIR_iter116/SVGLP/23.png}&
     \interpfig{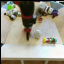}&
     \interpfig{Video_pred/BAIR_iter116/SVGLP/29.png}\\
       & SAVP Det&
     \interpfig{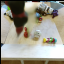}&
     \interpfig{Video_pred/BAIR_iter116/svap_det/7.png}&
     \interpfig{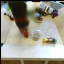}&
     \interpfig{Video_pred/BAIR_iter116/svap_det/15.png}&
     \interpfig{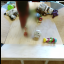}&
     \interpfig{Video_pred/BAIR_iter116/svap_det/23.png}&
     \interpfig{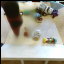}&
     \interpfig{Video_pred/BAIR_iter116/svap_det/29.png}\\ 
    & SAVP &
     \interpfig{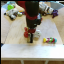}&
     \interpfig{Video_pred/BAIR_iter116/svap/7.png}&
     \interpfig{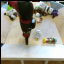}&
     \interpfig{Video_pred/BAIR_iter116/svap/15.png}&
     \interpfig{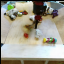}&
     \interpfig{Video_pred/BAIR_iter116/svap/23.png}&
     \interpfig{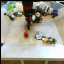}&
     \interpfig{Video_pred/BAIR_iter116/svap/29.png}\\ 
  \end{tabular}
  \caption{Qualitative results on the BAIR dataset (universally low-scoring sequence). All models had at least one time step for which their worst frame is from this sequence. Our method's failure mode differs from that of other methods. Other methods blur the moving object or previously occluded regions in the image. Our method consistently generates sharp images, but the visible objects are often subtly different from those in the ground truth frames.}
  \label{fig:worstbair}
  \end{figure}

\begin{figure}[h]
  \centering
  \setlength\tabcolsep{0.32pt}
  \renewcommand{\arraystretch}{0.12}
  \begin{tabular}{cccccccccc}
    &t=2  & t=4& t=8& t=12 &t=16 &t=20& t=24& t=28& t=30\\\\
     
     \rotatebox{90}{~~~~~~GT}&
     \interpfig{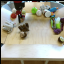}&
     \interpfig{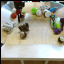}&
     \interpfig{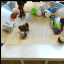}&
     \interpfig{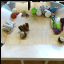}&
     \interpfig{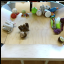}&
     \interpfig{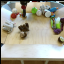}&
     \interpfig{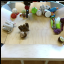}&
     \interpfig{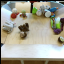}&
     \interpfig{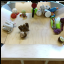}\\
    & Ours &
      \interpfig{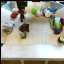}&
      \interpfig{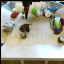}&
     \interpfig{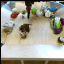}&
     \interpfig{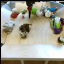}&
     \interpfig{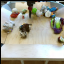}&
     \interpfig{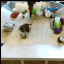}&
     \interpfig{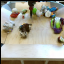}&
     \interpfig{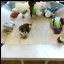}\\

     & SVGLP &
     \interpfig{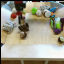}&
     \interpfig{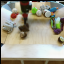}&
     \interpfig{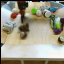}&
     \interpfig{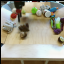}&
     \interpfig{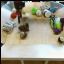}&
     \interpfig{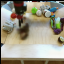}&
     \interpfig{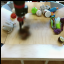}&
     \interpfig{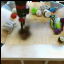}\\
    &  SAVP Det&
     \interpfig{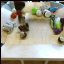}&
     \interpfig{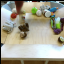}&
     \interpfig{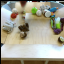}&
     \interpfig{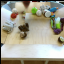}&
     \interpfig{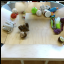}&
     \interpfig{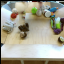}&
     \interpfig{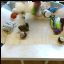}&
     \interpfig{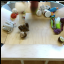}\\
    & SAVP &
     \interpfig{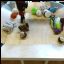}&
     \interpfig{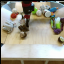}&
     \interpfig{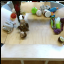}&
     \interpfig{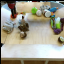}&
     \interpfig{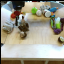}&
     \interpfig{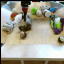}&
     \interpfig{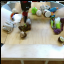}&
     \interpfig{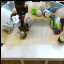}\\ 

     \\ \\ \rotatebox{90}{~~~~~~GT}&
     \interpfig{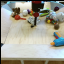}&
     \interpfig{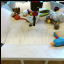}&
     \interpfig{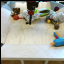}&
     \interpfig{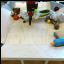}&
     \interpfig{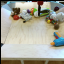}&
     \interpfig{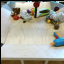}&
     \interpfig{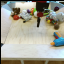}&
     \interpfig{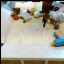}&
     \interpfig{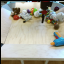}\\
     & Ours &
      \interpfig{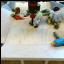}&
      \interpfig{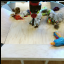}&
     \interpfig{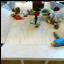}&
     \interpfig{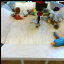}&
     \interpfig{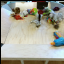}&
     \interpfig{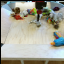}&
     \interpfig{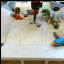}&
     \interpfig{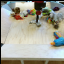}\\
     &  SVGLP&
     \interpfig{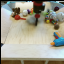}&
     \interpfig{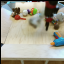}&
     \interpfig{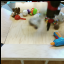}&
     \interpfig{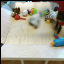}&
     \interpfig{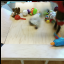}&
     \interpfig{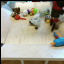}&
     \interpfig{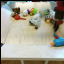}&
     \interpfig{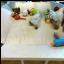}\\ & 
     SAVP Det&
     \interpfig{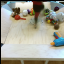}&
     \interpfig{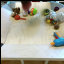}&
     \interpfig{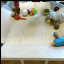}&
     \interpfig{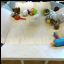}&
     \interpfig{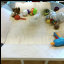}&
     \interpfig{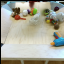}&
     \interpfig{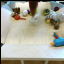}&
     \interpfig{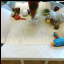}\\&
     SAVP&
     \interpfig{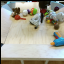}&
     \interpfig{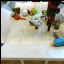}&
     \interpfig{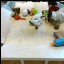}&
     \interpfig{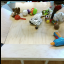}&
     \interpfig{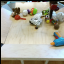}&
     \interpfig{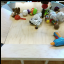}&
     \interpfig{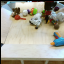}&
     \interpfig{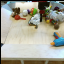}

  \end{tabular}
  \caption{Additional qualitative results for video prediction on the BAIR test set. The LSTM is conditioned on 2 images to generate the next 28.}
  \label{fig:bair_video_app2}
\end{figure}

\clearpage

\newpage
\section{KTH Dataset - Additional Video Prediction}

\begin{figure}[h]
  \centering
  \setlength\tabcolsep{0.32pt}
  \renewcommand{\arraystretch}{0.15}
  \begin{tabular}{cccccccccc}
    &t=11  &t=14 &t=17 &t=20 &t=23 &t=26 &t=29 &t=32 &t=35\\\\
    \rotatebox{90}{~~~GT}&
     \interpfig{Video_pred/KTH_iter181/GT/3.png}&
     \interpfig{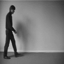}&
     \interpfig{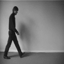}&
     \interpfig{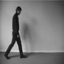}&
     \interpfig{Video_pred/KTH_iter181/GT/15.png}&
     \interpfig{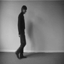}&
     \interpfig{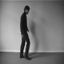}&
     \interpfig{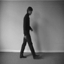}&
     \interpfig{Video_pred/KTH_iter181/GT/27.png}\\
        \rotatebox{90}{~~~Ours}&
     \interpfig{Video_pred/KTH_iter181/Ours_64/000009.png}&
     \interpfig{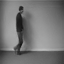}&
     \interpfig{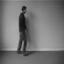}&
     \interpfig{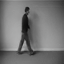}&
     \interpfig{Video_pred/KTH_iter181/Ours_64/000021.png}&
     \interpfig{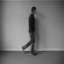}&
     \interpfig{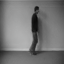}&
     \interpfig{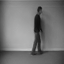}&
     \interpfig{Video_pred/KTH_iter181/Ours_64/000033.png}\\
    \rotatebox{90}{~~Savp Det}&
     \interpfig{Video_pred/KTH_iter181/SAVP_det/3.png}&
     \interpfig{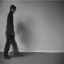}&
     \interpfig{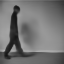}&
     \interpfig{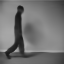}&
     \interpfig{Video_pred/KTH_iter181/SAVP_det/15.png}&
     \interpfig{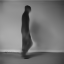}&
     \interpfig{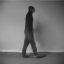}&
     \interpfig{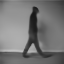}&
     \interpfig{Video_pred/KTH_iter181/SAVP_det/27.png}\\
   \rotatebox{90}{~~~Savp}&
     \interpfig{Video_pred/KTH_iter181/SAVP/3.png}&
     \interpfig{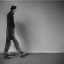}&
     \interpfig{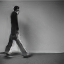}&
     \interpfig{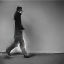}&
     \interpfig{Video_pred/KTH_iter181/SAVP/15.png}&
     \interpfig{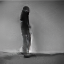}&
     \interpfig{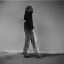}&
     \interpfig{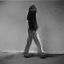}&
     \interpfig{Video_pred/KTH_iter181/SAVP/27.png}\\
   \rotatebox{90}{DRNet}&
     \interpfig{Video_pred/KTH_iter181/DRNET/0.png}&
     \interpfig{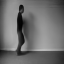}&
     \interpfig{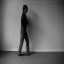}&
     \interpfig{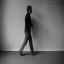}&
     \interpfig{Video_pred/KTH_iter181/DRNET/12.png}&
     \interpfig{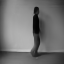}&
     \interpfig{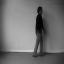}&
     \interpfig{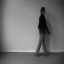}&
     \interpfig{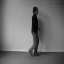}\\
     \end{tabular}
     \caption{Additional qualitative results from KTH action dataset.}
\end{figure}

\begin{figure}[h]
  \centering
  \setlength\tabcolsep{0.32pt}
  \renewcommand{\arraystretch}{0.15}
  \begin{tabular}{cccccccccc}
    &t=11  &t=14 &t=17 &t=20 &t=23 &t=26 &t=29 &t=32 &t=35\\\\
    \rotatebox{90}{~~~GT}&
     \interpfig{Video_pred/KTH_iter132/GT/3.png}&
     \interpfig{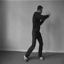}&
     \interpfig{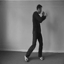}&
     \interpfig{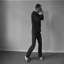}&
     \interpfig{Video_pred/KTH_iter132/GT/15.png}&
     \interpfig{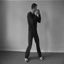}&
     \interpfig{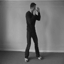}&
     \interpfig{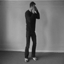}&
     \interpfig{Video_pred/KTH_iter132/GT/27.png}\\   \rotatebox{90}{~~~Ours}&
     \interpfig{Video_pred/KTH_iter132/Ours_64/000009.png}&
     \interpfig{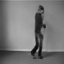}&
     \interpfig{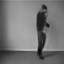}&
     \interpfig{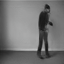}&
     \interpfig{Video_pred/KTH_iter132/Ours_64/000021.png}&
     \interpfig{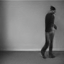}&
     \interpfig{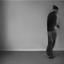}&
     \interpfig{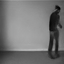}&
     \interpfig{Video_pred/KTH_iter132/Ours_64/000033.png}\\
    \rotatebox{90}{~~Savp Det}&
     \interpfig{Video_pred/KTH_iter132/SAVP_det/3.png}&
     \interpfig{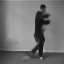}&
     \interpfig{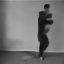}&
     \interpfig{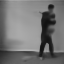}&
     \interpfig{Video_pred/KTH_iter132/SAVP_det/15.png}&
     \interpfig{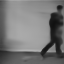}&
     \interpfig{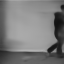}&
     \interpfig{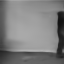}&
     \interpfig{Video_pred/KTH_iter132/SAVP_det/27.png}\\
   \rotatebox{90}{~~~Savp}&
     \interpfig{Video_pred/KTH_iter132/SAVP/3.png}&
     \interpfig{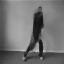}&
     \interpfig{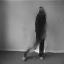}&
     \interpfig{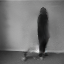}&
     \interpfig{Video_pred/KTH_iter132/SAVP/15.png}&
     \interpfig{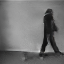}&
     \interpfig{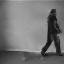}&
     \interpfig{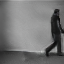}&
     \interpfig{Video_pred/KTH_iter132/SAVP/27.png}\\
        \rotatebox{90}{DRNet}&
     \interpfig{Video_pred/KTH_iter132/DRNET/0.png}&
     \interpfig{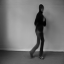}&
     \interpfig{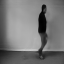}&
     \interpfig{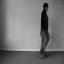}&
     \interpfig{Video_pred/KTH_iter132/DRNET/12.png}&
     \interpfig{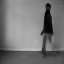}&
     \interpfig{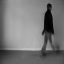}&
     \interpfig{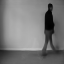}&
     \interpfig{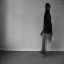}\\

     \end{tabular}
     \caption{Additional frames from one of the worst sequence shared by both the Savp and their Deterministic baseline model. One of the reasons the methods perform poorly in LPIPS metric is because the action is misinterpreted by all three models. All models eventually transition to predicting a walking person for a boxing sequence. Furthermore, notice how the structural integrity of the foreground object collapses starting at roughly frame 17 for Savp and Savp-Det. }
\end{figure}

\begin{figure}[h]
  \centering
  \setlength\tabcolsep{0.32pt}
  \renewcommand{\arraystretch}{0.15}
  \begin{tabular}{cccccccccc}
    &t=11  &t=14 &t=17 &t=20 &t=23 &t=26 &t=29 &t=32 &t=35\\\\
    \rotatebox{90}{~~~GT}&
     \interpfig{Video_pred/KTH_iter58/GT/3.png}&
     \interpfig{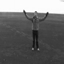}&
     \interpfig{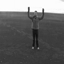}&
     \interpfig{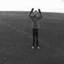}&
     \interpfig{Video_pred/KTH_iter58/GT/15.png}&
     \interpfig{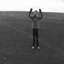}&
     \interpfig{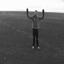}&
     \interpfig{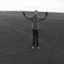}&
     \interpfig{Video_pred/KTH_iter58/GT/27.png}\\
    \rotatebox{90}{~~~Ours}&
     \interpfig{Video_pred/KTH_iter58/Ours_64/000009.png}&
     \interpfig{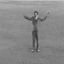}&
     \interpfig{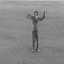}&
     \interpfig{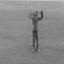}&
     \interpfig{Video_pred/KTH_iter58/Ours_64/000021.png}&
     \interpfig{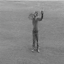}&
     \interpfig{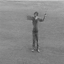}&
     \interpfig{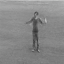}&
     \interpfig{Video_pred/KTH_iter58/Ours_64/000033.png}\\
    \rotatebox{90}{~~Savp Det}&
     \interpfig{Video_pred/KTH_iter58/SAVP_det/3.png}&
     \interpfig{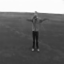}&
     \interpfig{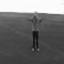}&
     \interpfig{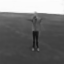}&
     \interpfig{Video_pred/KTH_iter58/SAVP_det/15.png}&
     \interpfig{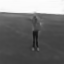}&
     \interpfig{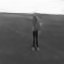}&
     \interpfig{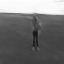}&
     \interpfig{Video_pred/KTH_iter58/SAVP_det/27.png}\\
   \rotatebox{90}{~~~Savp}&
     \interpfig{Video_pred/KTH_iter58/SAVP/3.png}&
     \interpfig{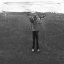}&
     \interpfig{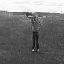}&
     \interpfig{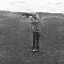}&
     \interpfig{Video_pred/KTH_iter58/SAVP/15.png}&
     \interpfig{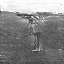}&
     \interpfig{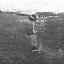}&
     \interpfig{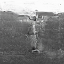}&
     \interpfig{Video_pred/KTH_iter58/SAVP/27.png}\\
   \rotatebox{90}{DRNet}&
     \interpfig{Video_pred/KTH_iter58/DRNET/0.png}&
     \interpfig{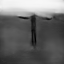}&
     \interpfig{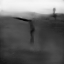}&
     \interpfig{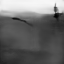}&
     \interpfig{Video_pred/KTH_iter58/DRNET/12.png}&
     \interpfig{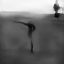}&
     \interpfig{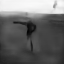}&
     \interpfig{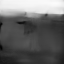}&
     \interpfig{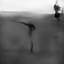}\\

     \end{tabular}
     \caption{Qualitative results from one of our worst performing KTH sequence. Interestingly, all the methods fail in this sequence. Our method performs badly because it does not able to reconstruct the novel background and person's appearance, even though it is able to generate poses. On the other hand, SAVP's deterministic baseline  and SAVP stochastic models are able to reconstruct the images for few frames thanks to the skip connections between the time frames but the pose and appearance blur roughly after 15 frames. DRNet does a slightly better job at preserving the background than our method, though their foreground loses structural integrity very quickly as well.}
\end{figure}

%% file: 0-main.bbl
\begin{thebibliography}{10}\itemsep=-1pt

\bibitem{babaeizadeh2017stochastic}
M.~Babaeizadeh, C.~Finn, D.~Erhan, R.~H. Campbell, and S.~Levine.
\newblock Stochastic variational video prediction.
\newblock {\em arXiv preprint arXiv:1710.11252}, 2017.

\bibitem{Charles13}
J.~Charles, T.~Pfister, D.~Magee, D.~Hogg, and A.~Zisserman.
\newblock Domain adaptation for upper body pose tracking in signed {TV}
  broadcasts.
\newblock In {\em British Machine Vision Conference}, 2013.

\bibitem{chen2019optimal}
Y.~Chen, T.~T. Georgiou, and A.~Tannenbaum.
\newblock Optimal transport for gaussian mixture models.
\newblock {\em IEEE Access}, 7:6269--6278, 2019.

\bibitem{denton2018stochastic}
E.~Denton and R.~Fergus.
\newblock Stochastic video generation with a learned prior.
\newblock {\em arXiv preprint arXiv:1802.07687}, 2018.

\bibitem{denton2017unsupervised}
E.~L. Denton et~al.
\newblock Unsupervised learning of disentangled representations from video.
\newblock In {\em Advances in neural information processing systems}, pages
  4414--4423, 2017.

\bibitem{dinh2016density}
L.~Dinh, J.~Sohl-Dickstein, and S.~Bengio.
\newblock Density estimation using real nvp.
\newblock {\em arXiv preprint arXiv:1605.08803}, 2016.

\bibitem{finn2016unsupervised}
C.~Finn, I.~Goodfellow, and S.~Levine.
\newblock Unsupervised learning for physical interaction through video
  prediction.
\newblock In {\em Advances in neural information processing systems}, pages
  64--72, 2016.

\bibitem{hochreiter1997lstm}
S.~Hochreiter and J.~Schmidhuber.
\newblock Lstm can solve hard long time lag problems.
\newblock In {\em Advances in neural information processing systems}, pages
  473--479, 1997.

\bibitem{jakab2018unsupervised}
T.~Jakab, A.~Gupta, H.~Bilen, and A.~Vedaldi.
\newblock Unsupervised learning of object landmarks through conditional image
  generation.
\newblock In {\em Advances in Neural Information Processing Systems}, 2018.

\bibitem{jiang2018super}
H.~Jiang, D.~Sun, V.~Jampani, M.-H. Yang, E.~Learned-Miller, and J.~Kautz.
\newblock Super slomo: High quality estimation of multiple intermediate frames
  for video interpolation.
\newblock In {\em Proceedings of the IEEE Conference on Computer Vision and
  Pattern Recognition}, pages 9000--9008, 2018.

\bibitem{Kanazawa_2016_CVPR}
A.~Kanazawa, D.~W. Jacobs, and M.~Chandraker.
\newblock Warpnet: Weakly supervised matching for single-view reconstruction.
\newblock In {\em The IEEE Conference on Computer Vision and Pattern
  Recognition (CVPR)}, June 2016.

\bibitem{kingma2018glow}
D.~P. Kingma and P.~Dhariwal.
\newblock Glow: Generative flow with invertible 1x1 convolutions.
\newblock In {\em Advances in Neural Information Processing Systems}, pages
  10215--10224, 2018.

\bibitem{kingma2013auto}
D.~P. Kingma and M.~Welling.
\newblock Auto-encoding variational bayes.
\newblock {\em arXiv preprint arXiv:1312.6114}, 2013.

\bibitem{kumar2019videoflow}
M.~Kumar, M.~Babaeizadeh, D.~Erhan, C.~Finn, S.~Levine, L.~Dinh, and D.~Kingma.
\newblock Videoflow: A flow-based generative model for video.
\newblock {\em arXiv preprint arXiv:1903.01434}, 2019.

\bibitem{laptev2004recognizing}
I.~Laptev, B.~Caputo, et~al.
\newblock Recognizing human actions: a local svm approach.
\newblock pages 32--36. IEEE, 2004.

\bibitem{lee2018stochastic}
A.~X. Lee, R.~Zhang, F.~Ebert, P.~Abbeel, C.~Finn, and S.~Levine.
\newblock Stochastic adversarial video prediction.
\newblock {\em arXiv preprint arXiv:1804.01523}, 2018.

\bibitem{liu2017video}
Z.~Liu, R.~A. Yeh, X.~Tang, Y.~Liu, and A.~Agarwala.
\newblock Video frame synthesis using deep voxel flow.
\newblock In {\em Proceedings of the IEEE International Conference on Computer
  Vision}, 2017.

\bibitem{lorenz2019unsupervised}
D.~Lorenz, L.~Bereska, T.~Milbich, and B.~Ommer.
\newblock Unsupervised part-based disentangling of object shape and appearance.
\newblock In {\em CVPR}, 2019.

\bibitem{madyasthaekf}
V.~Madyastha, V.~Ravindra, S.~Mallikarjunan, and A.~Goyal.
\newblock Extended kalman filter vs. error state kalman filter for aircraft
  attitude estimation.
\newblock 08 2011.

\bibitem{mathieu2015deep}
M.~Mathieu, C.~Couprie, and Y.~LeCun.
\newblock Deep multi-scale video prediction beyond mean square error.
\newblock {\em arXiv preprint arXiv:1511.05440}, 2015.

\bibitem{miyato2018spectral}
T.~Miyato, T.~Kataoka, M.~Koyama, and Y.~Yoshida.
\newblock Spectral normalization for generative adversarial networks.
\newblock In {\em International Conference on Learning Representations (ICLR)},
  2018.

\bibitem{AvinashSlomo}
A.~Paliwal.
\newblock Super-slomo.
\newblock \url{https://github.com/avinashpaliwal/Super-SloMo}.

\bibitem{park2019semantic}
T.~Park, M.-Y. Liu, T.-C. Wang, and J.-Y. Zhu.
\newblock Semantic image synthesis with spatially-adaptive normalization.
\newblock In {\em CVPR}, 2019.

\bibitem{pfister2015flowing}
T.~Pfister, J.~Charles, and A.~Zisserman.
\newblock Flowing convnets for human pose estimation in videos.
\newblock In {\em Proceedings of the IEEE International Conference on Computer
  Vision}, pages 1913--1921, 2015.

\bibitem{pottorff2019video}
R.~Pottorff, J.~Nielsen, and D.~Wingate.
\newblock Video extrapolation with an invertible linear embedding.
\newblock {\em arXiv preprint arXiv:1903.00133}, 2019.

\bibitem{reda2018sdc}
F.~A. Reda, G.~Liu, K.~J. Shih, R.~Kirby, J.~Barker, D.~Tarjan, A.~Tao, and
  B.~Catanzaro.
\newblock Sdc-net: Video prediction using spatially-displaced convolution.
\newblock In {\em Proceedings of the European Conference on Computer Vision
  (ECCV)}, pages 718--733, 2018.

\bibitem{reda2019unsupervised}
F.~A. Reda, D.~Sun, A.~Dundar, M.~Shoeybi, G.~Liu, K.~J. Shih, A.~Tao,
  J.~Kautz, and B.~Catanzaro.
\newblock Unsupervised video interpolation using cycle consistency.
\newblock {\em arXiv preprint arXiv:1906.05928}, 2019.

\bibitem{reed2015deep}
S.~E. Reed, Y.~Zhang, Y.~Zhang, and H.~Lee.
\newblock Deep visual analogy-making.
\newblock In {\em Advances in neural information processing systems}, pages
  1252--1260, 2015.

\bibitem{ronneberger2015u}
O.~Ronneberger, P.~Fischer, and T.~Brox.
\newblock U-net: Convolutional networks for biomedical image segmentation.
\newblock In {\em International Conference on Medical image computing and
  computer-assisted intervention}, 2015.

\bibitem{schuldt2004recognizing}
C.~Schuldt, I.~Laptev, and B.~Caputo.
\newblock Recognizing human actions: a local svm approach.
\newblock In {\em Proceedings of the 17th International Conference on Pattern
  Recognition, 2004. ICPR 2004.}, volume~3, pages 32--36. IEEE, 2004.

\bibitem{siarohin2018animating}
A.~Siarohin, S.~Lathuili{\`e}re, S.~Tulyakov, E.~Ricci, and N.~Sebe.
\newblock Animating arbitrary objects via deep motion transfer.
\newblock {\em arXiv preprint arXiv:1812.08861}, 2018.

\bibitem{suwajanakorn2018discovery}
S.~Suwajanakorn, N.~Snavely, J.~J. Tompson, and M.~Norouzi.
\newblock Discovery of latent 3d keypoints via end-to-end geometric reasoning.
\newblock In {\em Advances in Neural Information Processing Systems}, pages
  2059--2070, 2018.

\bibitem{thewlis2017unsupervised}
J.~Thewlis, H.~Bilen, and A.~Vedaldi.
\newblock Unsupervised learning of object frames by dense equivariant image
  labelling.
\newblock In {\em Advances in Neural Information Processing Systems}, pages
  844--855, 2017.

\bibitem{Thewlis17}
J.~Thewlis, H.~Bilen, and A.~Vedaldi.
\newblock Unsupervised learning of object landmarks by factorized spatial
  embeddings.
\newblock In {\em International Conference on Computer Vision (ICCV)}, 2017.

\bibitem{tulyakov2018mocogan}
S.~Tulyakov, M.-Y. Liu, X.~Yang, and J.~Kautz.
\newblock Mocogan: Decomposing motion and content for video generation.
\newblock In {\em Proceedings of the IEEE conference on computer vision and
  pattern recognition}, pages 1526--1535, 2018.

\bibitem{unterthiner2018towards}
T.~Unterthiner, S.~van Steenkiste, K.~Kurach, R.~Marinier, M.~Michalski, and
  S.~Gelly.
\newblock Towards accurate generative models of video: A new metric \&
  challenges.
\newblock {\em arXiv preprint arXiv:1812.01717}, 2018.

\bibitem{villegas2017learning}
R.~Villegas, J.~Yang, Y.~Zou, S.~Sohn, X.~Lin, and H.~Lee.
\newblock Learning to generate long-term future via hierarchical prediction.
\newblock In {\em Proceedings of the 34th International Conference on Machine
  Learning-Volume 70}, pages 3560--3569. JMLR. org, 2017.

\bibitem{vondrick2017generating}
C.~Vondrick and A.~Torralba.
\newblock Generating the future with adversarial transformers.
\newblock In {\em Proceedings of the IEEE Conference on Computer Vision and
  Pattern Recognition}, pages 1020--1028, 2017.

\bibitem{walker2015dense}
J.~Walker, A.~Gupta, and M.~Hebert.
\newblock Dense optical flow prediction from a static image.
\newblock In {\em Proceedings of the IEEE International Conference on Computer
  Vision}, 2015.

\bibitem{wichers2018hierarchical}
N.~Wichers, R.~Villegas, D.~Erhan, and H.~Lee.
\newblock Hierarchical long-term video prediction without supervision.
\newblock {\em arXiv preprint arXiv:1806.04768}, 2018.

\bibitem{xue2016visual}
T.~Xue, J.~Wu, K.~Bouman, and B.~Freeman.
\newblock Visual dynamics: Probabilistic future frame synthesis via cross
  convolutional networks.
\newblock In {\em Advances in Neural Information Processing Systems}, 2016.

\bibitem{LPIPSGithub}
R.~Zhang.
\newblock https://github.com/richzhang/perceptualsimilarity.
\newblock \url{https://github.com/richzhang/PerceptualSimilarity}.

\bibitem{zhang2018perceptual}
R.~Zhang, P.~Isola, A.~A. Efros, E.~Shechtman, and O.~Wang.
\newblock The unreasonable effectiveness of deep features as a perceptual
  metric.
\newblock In {\em CVPR}, 2018.

\bibitem{Zhang_2018_CVPR}
Y.~Zhang, Y.~Guo, Y.~Jin, Y.~Luo, Z.~He, and H.~Lee.
\newblock Unsupervised discovery of object landmarks as structural
  representations.
\newblock In {\em The IEEE Conference on Computer Vision and Pattern
  Recognition (CVPR)}, June 2018.

\end{thebibliography}
